\documentclass[11pt]{article}

\usepackage[preprint]{acl}

\usepackage{times}
\usepackage{latexsym}
\usepackage[T1]{fontenc}
\usepackage[utf8]{inputenc}
\usepackage{microtype}
\usepackage{inconsolata}
\usepackage{graphicx}
\usepackage{amsmath}
\usepackage{amssymb}
\usepackage{booktabs}
\usepackage{array}
\usepackage{xcolor}

\newcommand{\method}{SBBT}
\newcommand{\Prob}{\mathbb{P}}
\newcommand{\stateon}{\ensuremath{\mathrm{H}}}
\newcommand{\stateoff}{\ensuremath{\mathrm{L}}}
\newcommand{\Brier}{\operatorname{Brier}}
\newcommand{\AUROC}{\operatorname{AUROC}}
\newcommand{\Bin}{\operatorname{bin}}

\newcommand{\calF}{\mathcal{F}}
\newcommand{\Thetapre}{\Theta_{\mathrm{pre}}}
\newcommand{\PFC}{\mathrm{PFC}}
\newcommand{\Std}{\mathrm{std}}
\newcommand{\Audit}{\mathrm{audit}}

\title{Prefix-Safe Bayesian Belief Tracking for LLM Reasoning Reliability: Separating Calibration from Ranking}

\author{
Zhenghan Song \\
Cornell University \\
\texttt{zs448@cornell.edu}
\And
Yunyi Li \\
Columbia University \\
\texttt{yl5652@columbia.edu}
\And
Yulong Liu\thanks{Corresponding author.} \\
Cornell University \\
\texttt{yl3825@cornell.edu}
}

\begin{document}
\maketitle

\begin{abstract}
Long reasoning traces need reliability estimates before final answers are known. We study prefix-conditioned eventual-success estimation, $\Prob(y{=}1\,\vert\,o_{1:t})$, using prefix-safe observations. Sequential Bayesian Belief Tracking (\method) calibrates observation likelihoods and recursively updates a two-state belief, providing a common tracker for scalar scores, text and self-verification markers, hidden clusters, token-pooling probes, and latent-trajectory features. Across generated open-weight traces on MATH-500, GSM8K, AIME 2025, and RIMO-N, probability quality and ranking separate: score-only \method{} often improves Brier, while AUROC gains require structure-aware evidence beyond strong prefix-safe baselines. In the strongest hard math setting, structure-aware observations reach $+0.110$ AUROC against standard prefix-safe baselines. Under a same-prefix classifier audit, MATH-500 text markers and RIMO-N self-verification signals remain positive. Together, these findings support \method{} as a calibration-aware online inference framework and expose an evidence regime: scalar scores mainly support probability quality, while structure-aware prefix signals support ranking only when strong prefix-safe baselines have not already absorbed the rank evidence.
\end{abstract}

\noindent{\scriptsize\textbf{Code:} \href{https://github.com/HaningZS/Bayesian-Belief-Tracking.git}{\texttt{github.com/HaningZS/Bayesian-Belief-Tracking}}}

\section{Introduction}

Reliability estimation for long reasoning traces is an online belief-updating problem under prefix-safety constraints: while generation is unfolding, a monitor must decide whether the observed prefix still supports eventual success. The practical question is not only whether a sampled final answer is correct, but whether the observed prefix still provides evidence of eventual success before that answer is known. Intervention, early stopping, and review triage all need prefix-time reliability estimates that can be computed during generation, not after final correctness is available. This connects to selective prediction and confidence-based early stopping \citep{geifman2019selectivenet,schuster2022confident}, and to compute-aware inference \citep{chen2023frugalgpt,yang2025dynamic,eisenstadt2025overclocking}.

The central difficulty is that prefix evidence is noisy and family-dependent. Calibration work shows that model probabilities and verbalized confidence are not reliable by default \citep{guo2017calibration,desai2020calibration,kadavath2022language}. Entropy, consistency, and semantic-uncertainty signals remain distribution-dependent \citep{lin2022teaching,tian2023just,farquhar2024detecting}. Prefix-level calibration is also setting-dependent \citep{xie2024calibrating}. In our setting, token entropy, verifier confidence, hidden-state probes, text markers such as self-correction, and latent-trajectory features can all correlate with final correctness, but their meaning changes across models, datasets, trace lengths, and prompt protocols. Empirically, scalar scores often improve probability quality without improving AUROC, while hidden-cluster observations or self-verification markers can help long-trace ranking yet fail against length-aware AIME baselines. The challenge is to identify non-redundant prefix-safe evidence and accumulate it over time.

We frame reasoning reliability as prefix-conditioned eventual-success belief tracking. Given observations $o_{1:t}$ extracted from a reasoning prefix, the target is $\Prob(y{=}1\,\vert\,o_{1:t})$, where $y$ is final-answer correctness. Propagated final labels provide weak supervision for eventual-success reliability states. Process-error localization requires separate process-labeled validation.

Sequential Bayesian Belief Tracking (\method) is a calibration-aware online belief tracker for this setting. \method{} first calibrates observation likelihoods on question-level train/calibration splits, then applies a two-state Bayesian filter to produce an online belief trajectory. The observation interface accepts continuous score observations, discrete concept codes, hidden-cluster assignments, self-verification markers, token-pooling probes, and latent-trajectory scores. This gives a controlled way to hold temporal integration fixed while changing the prefix signal under the same prefix-safe protocol \citep{rabiner1989tutorial,sarkka2013bayesian,guo2017calibration}.

Experiments expose this evidence regime. On MATH-500, DeepSeek-based rows give gains for structure-aware observations, including text patterns, hidden-cluster observations, and latent-trajectory features. GSM8K provides an easier cross-dataset check. RIMO-N tests whether hidden-cluster and self-verification signals transfer across 14B protocols under matched prompt settings. AIME and Qwen-family MATH rows show where learned-prefix or length-aware baselines dominate. These stress tests keep score-only filtering framed as a probability-quality tool.

The contribution is a task formulation, a tracker, and an evidence decomposition. We define prefix-conditioned eventual-success estimation as an online reliability inference task for long reasoning traces, separating prefix-safe monitoring from offline process-error localization. We introduce \method{} as a calibration-aware Bayesian tracker that converts heterogeneous prefix-safe observations into online belief trajectories under a common sequential update. We then decompose when scalar confidence, text and self-verification markers, hidden clusters, and trajectory features support probability quality or ranking. Appendix~\ref{sec:claim-evidence} summarizes the claim/evidence alignment.

\section{Related Work}

\paragraph{Reasoning traces, benchmarks, and supervision.}
Chain-of-thought prompting and self-consistency make intermediate reasoning explicit and improve mathematical reasoning by sampling or aggregating solution paths \citep{wei2022chain,wang2023selfconsistency}. GSM8K, MATH, and RIMO-N provide final-answer evaluation settings for arithmetic, competition-style, and olympiad-level reasoning \citep{cobbe2021training,hendrycks2021measuring,chen2025rimo}. Process-supervised reward models and process benchmarks use step-level or first-error annotations to evaluate intermediate reasoning quality \citep{uesato2022solving,lightman2023verify,wang2024mathshepherd}. Automated process-supervision and process-benchmark work provide adjacent validation settings \citep{zheng2024processbench,luo2024improve}. Our setting uses weaker but scalable final-answer supervision: it estimates prefix-conditioned eventual success, not step correctness.

\paragraph{Online monitoring and compute-aware inference.}
Selective prediction, adaptive language modeling, and compute-aware inference study when a system should abstain, stop, route, or allocate more computation under uncertainty \citep{geifman2019selectivenet,schuster2022confident,chen2023frugalgpt}. Reasoning-specific work studies early exit and reasoning-length control for long chain-of-thought generation \citep{yang2025dynamic,eisenstadt2025overclocking}. These settings motivate reliability estimates before the final answer is known. \method{} targets single-trace prefix-safe monitoring: at prefix $t$, the monitor may use only current-prefix observations and pre-fit calibration parameters, not multiple completed samples or future trace tokens.

\paragraph{Prefix evidence for reliability.}
Prior work studies uncertainty signals such as model confidence, verbalized uncertainty, semantic entropy, and internal consistency for detecting when a model is likely to be correct or unreliable \citep{kadavath2022language,lin2022teaching,tian2023just}. Semantic-entropy and internal-consistency work connect these signals to generated reasoning and open-ended answers \citep{farquhar2024detecting,xie2024calibrating}. A complementary line uses internal representations. Probing work shows that hidden states can encode task-relevant or truth-related information \citep{alain2016understanding,belinkov2022probing,burns2023discovering}. Reasoning-specific studies find that hidden states, self-verification probes, probe trajectories, and latent trajectories can predict future answer correctness or productive reasoning paths \citep{zhang2025reasoningknow,vilas2025tracing,chrabaszcz2026monitoring}. In \method{}, these signals are observation families consumed by the same calibration-aware sequential tracker, not standalone detectors.

\paragraph{Calibration and sequential probabilistic tracking.}
Calibration work motivates probability-quality metrics such as Brier score and expected calibration error \citep{brier1950verification,guo2017calibration}. Hidden Markov models and Bayesian filtering provide classical tools for integrating noisy observations over time \citep{rabiner1989tutorial,sarkka2013bayesian}. \method{} uses these tools for prefix-safe reliability tracking, with likelihood calibration, online filtering, and offline smoothing diagnostics separated.

\section{Task and Notation}

A reasoning trace consists of a question $q$, a generated sequence segmented into $T$ prefixes, observations $o_{1:T}$, and a final-answer label $y \in \{0,1\}$. The label is obtained by final-answer checking, with $y=1$ denoting an eventually correct trace. At prefix $t$, an online method may use only $o_{1:t}$ and metadata available before or at that prefix. The online target is
\begin{equation}
 b_t = \Prob(y{=}1\,\vert\,o_{1:t}).
\end{equation}
This target differs from step-level correctness. A prefix can look reliable under traces that eventually succeed even if no external annotation identifies the first mathematically wrong step. Process-error claims are reserved for future validation with process-labeled data.

We evaluate methods at the trace level by using the final online score from each trace unless otherwise specified. We also evaluate the belief at fixed prefix fractions such as 5\%, 50\%, and 100\%. Splits are question-level, so traces from the same question cannot cross train, calibration, and test partitions.

\section{Method}

\subsection{Overview}

\begin{figure*}[t]
 \centering
 \includegraphics[width=0.72\textwidth]{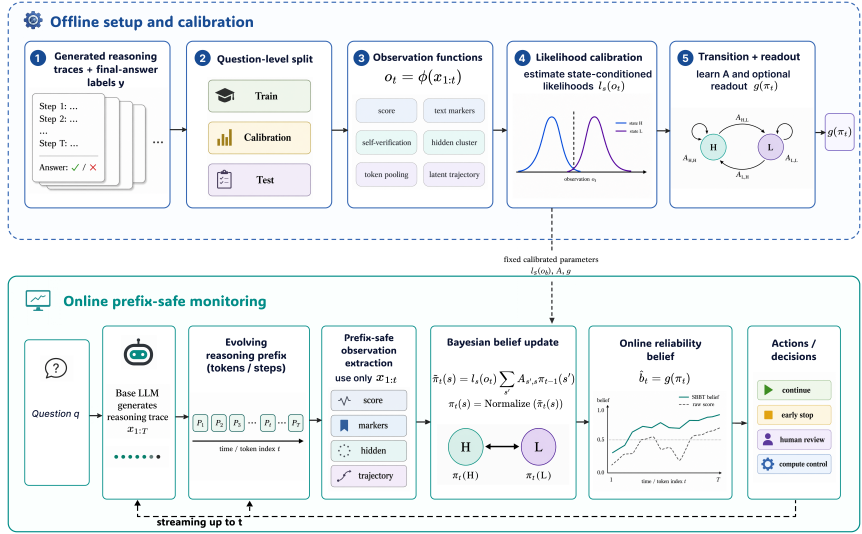}
 \caption{Prefix-safe belief tracking workflow. Offline calibration fits observation functions, likelihoods, transition dynamics, and an optional readout. Online monitoring maps each prefix $x_{1:t}$ to $o_t$ and updates an eventual-success belief.}
 \label{fig:pipeline}
\end{figure*}

\method{} maps prefix observations through observation extraction, likelihood calibration, and Bayesian filtering. Each prefix yields a score, a concept code, or both. Calibration fits state-conditioned likelihoods on held-out traces, and the filter combines those likelihoods with transition dynamics. Figure~\ref{fig:pipeline} summarizes the offline and online workflow.

\subsection{Observation Families}

An observation function $\phi$ maps $x_{1:t}$ to $o_t$. We evaluate score, text/self-verification, hidden-cluster observations, token-pooling, and latent-trajectory families under the same prefix-safe interface. Hybrid and joint variants combine score and concept likelihoods. Table~\ref{tab:observation-extraction} gives the fitting boundary and test-time information.

\begin{table*}[t]
\centering
\scriptsize
\setlength{\tabcolsep}{2pt}
\renewcommand{\arraystretch}{1.08}
\resizebox{0.88\linewidth}{!}{%
\begin{tabular}{@{}>{\raggedright\arraybackslash}p{0.16\linewidth}>{\raggedright\arraybackslash}p{0.16\linewidth}>{\raggedright\arraybackslash}p{0.24\linewidth}>{\raggedright\arraybackslash}p{0.16\linewidth}>{\raggedright\arraybackslash}p{0.16\linewidth}>{\raggedright\arraybackslash}p{0.10\linewidth}@{}}
\toprule
Observation family & Prefix input & Feature definition & Fitted on & Test-time information & Settings \\
\midrule
Score-only observations & Prefix score source through $x_{1:t}$ & Scalar score $a_t$ from entropy, verifier/probe probability, or trajectory-derived score. & Train/calibration questions for orientation and likelihood calibration. & Current prefix score only. & Score bins; calibration mode. \\
Text markers & Generated text prefix $x_{1:t}$ & Deterministic code for calculation, correction, uncertainty, and final-answer language. & Rules fixed before held-out evaluation. & Current prefix text only. & Marker lexicon. \\
Self-verification markers & Generated text prefix $x_{1:t}$ & Deterministic code for verification, self-correction, uncertainty, and answer commitment. & Rules fixed before held-out evaluation. & Current prefix text only. & Marker lexicon. \\
Hidden-cluster observations & Prefix hidden vector from $x_{1:t}$ & Train-split cluster id assigned to each held-out prefix vector. & Train questions only; calibration questions estimate emissions. & Current prefix hidden vector only. & Feature field; cluster count. \\
Hidden-probe and token-pooling scores & Prefix hidden vectors through $x_{1:t}$ & Probe probability from last-token, mean-pool, max-pool, or scoped pooled features. & Train questions for probe fitting; calibration questions for score calibration. & Prefix hidden features within the declared scope. & Pool scope; probe type. \\
Latent-trajectory / activation features & Hidden-vector sequence observed up to prefix $t$ & Net change, cumulative change, aligned change, or activation-trajectory scalar. & Train questions for score orientation; calibration questions for likelihoods. & Prefix trajectory up to $t$ only. & Metric; layer/feature field. \\
\bottomrule
\end{tabular}}
\caption{Observation extraction summary. Rows define prefix-safe observation families, their fitted components, and their test-time information.}
\label{tab:observation-extraction}
\end{table*}

This interface lets the same filtering core test whether each evidence family is informative under the prefix-safe boundary. Saturated scalar scores remain visible as a limitation, while hidden-cluster observations or self-verification markers can be swapped in without changing the tracker.

\subsection{Calibration}

\method{} estimates likelihoods $\Prob(o_t \mid z_t)$ for a latent reliability state $z_t \in \{\stateon,\stateoff\}$, where $\stateon$ denotes the high-reliability state and $\stateoff$ denotes the low-reliability state. In the default all-prefix calibration mode, final trace labels are propagated to prefixes as weak supervision for eventual success. We also evaluate final-step and EM calibration variants to avoid treating this weak-supervision choice as a hidden assumption.

For continuous scores, calibration uses histogram, quantile, or learned likelihood estimators over score values. For concept observations, calibration estimates smoothed categorical likelihoods. Hybrid filtering multiplies score and concept likelihoods under a conditional-independence approximation, while joint filtering estimates $\Prob(\text{score bin}, \text{concept} \mid z_t)$ directly. Since prefix observations are cumulative and highly autocorrelated, this is a misspecified monitoring model, not a generative claim about the trace text. Calibration is performed only on train/calibration questions, and test questions are never used to fit emissions, probes, orientations, clusters, or thresholds.

\subsection{Bayesian Filtering}

The online belief update is a two-state hidden Markov filter. Let $\pi_{t-1}(s)$ be the posterior belief for state $s$ after prefix $t-1$, $A_{s's}=\Prob(z_t=s \mid z_{t-1}=s')$ be the transition matrix, and let $\ell_s(o_t)$ be the calibrated emission likelihood value, a probability mass for discrete observations or a density value for continuous scores. The filter update is
\begin{align}
 \tilde{\pi}_t(s) &= \ell_s(o_t) \sum_{s'} A_{s's}\,\pi_{t-1}(s'), \\
 \pi_t(s) &= \frac{\tilde{\pi}_t(s)}{\sum_{s^\star}\tilde{\pi}_t(s^\star)}.
\end{align}
We report $\pi_t(\stateon)$ as the default model-based online reliability score. A calibrated target-probability readout can additionally map this posterior through a calibration function or an outcome-readout vector fitted on calibration prefixes. Appendix~\ref{sec:math-details} gives the formal distinction. Stationary transitions are the default. Non-stationary transitions are evaluated as an ablation. Viterbi decoding and forward-backward smoothing are used only as offline diagnostics because they depend on full-trace information.

\paragraph{Evidence view.}
In the two-state case, the filtering update also has a log-odds form. Let $\mu_t(s)=\sum_{s'}A_{s's}\pi_{t-1}(s')$ be the predictive belief before observing $o_t$. When all predictive masses and emission likelihoods are positive, the update implies
\begin{equation}
\log \frac{\pi_t(\stateon)}{\pi_t(\stateoff)}
=
\log \frac{\mu_t(\stateon)}{\mu_t(\stateoff)}
+
\log \frac{\ell_{\stateon}(o_t)}{\ell_{\stateoff}(o_t)}.
\end{equation}
Each prefix observation contributes a calibrated log-likelihood-ratio evidence term. This view keeps temporal integration fixed while changing the observation family: scalar scores, text markers, hidden clusters, and trajectory features differ by the evidence terms they provide to the same filter. Appendix~\ref{sec:log-odds-evidence} gives the derivation.

\section{Experiments}

\subsection{Setup}

We evaluate generated open-weight traces from GSM8K, MATH-500, AIME 2025 via MathArena \citep{matharena2025aime,dekoninck2026matharena}, and RIMO-N. Main MATH rows use DeepSeek-R1-Distill-Qwen models \citep{deepseekai2025r1}, including a full MATH-500 Level 5 DeepSeek-Qwen-7B run with 134 questions and four traces per question. Cross-model and stress-test rows use Qwen2.5-Math-7B-Instruct, Qwen3-8B, Qwen3-14B, DeepSeek-Qwen-14B, and an appendix DeepSeek-Llama-8B diagnostic \citep{yang2024qwen25math,yang2025qwen3,grattafiori2024llama3}. AIME and RIMO use longer token budgets and token-window segmentation to preserve prefix-safe hidden features. MATH-500 Level labels are inherited difficulty slices from the MATH benchmark and H4 MATH-500 release \citep{hendrycks2021measuring,huggingfaceh4math500}. These Levels 1 to 5 define problem subsets only. They are not \method{} variants, neural-network layers, or training stages.

All primary evaluations use question-level train/calibration/test splits and split-seed robustness sweeps. We report trace error rate, mean observations per trace, score saturation checks, AUROC gaps against the best prefix-safe baseline, positive-gap fraction across split seeds, and Brier deltas against calibrated temporal baselines. Additional appendix analyses cover Qwen3 verification-span scoped pooling and rollout belief joins. These rows are not used as main rank evidence.

We organize the evaluation around five checks: rank against the standard prefix-safe baseline set, Brier and calibration behavior, observation-family ablations, early-prefix and utility behavior, and rollout-based empirical-continuation validation. Together, these checks test whether an observation family contains prefix evidence that the baseline has not already captured.

\subsection{Baselines and Metrics}

Table~\ref{tab:main} reports AUROC gaps against standard prefix-safe baselines: score summaries, moving averages, calibrated score baselines, temporal summaries, and a deterministic learned score/length baseline. PFC is used separately as a same-prefix evidence-redundancy audit, not as part of the Table~\ref{tab:main} gaps. This keeps the standard-baseline evidence map and the stricter PFC audit as distinct tests. A positive post-PFC gap means that the evaluated \method{} variant beats both the standard baselines and PFC under the tested feature class. Appendix~\ref{sec:pfc-audit-algebra} gives the algebra. 

Table~\ref{tab:method-scope} makes this split explicit, separating primary headline baselines from audits and context methods.

\begin{table}[t]
\centering
\scriptsize
\setlength{\tabcolsep}{2pt}
\renewcommand{\arraystretch}{1.06}
\begin{tabular}{@{}>{\raggedright\arraybackslash}p{0.24\linewidth}>{\raggedright\arraybackslash}p{0.25\linewidth}>{\centering\arraybackslash}p{0.16\linewidth}>{\raggedright\arraybackslash}p{0.25\linewidth}@{}}
\toprule
Comparator group & Purpose & In Table~\ref{tab:main}? & Where reported \\
\midrule
Standard score/temporal baselines & Primary prefix-safe comparison & Yes & Table~\ref{tab:main}; Figures~\ref{fig:gaps}--\ref{fig:rank-calibration} \\
Learned score/length baseline & Strong standard prefix-safe baseline & Yes & Table~\ref{tab:main}; Figure~\ref{fig:rank-calibration} \\
Prefix-feature classifier (PFC) & Same-prefix evidence-redundancy audit & No & Appendix~\ref{sec:claim-evidence} \\
Multi-sample inference & Offline context & No & Related Work / discussion \\
Process-supervised methods & Step/process validation context & No & Related Work / limitations \\
\bottomrule
\end{tabular}
\caption{Comparator scope for primary results and audits. The ``In Table~\ref{tab:main}?'' column indicates whether a comparator group is included in the headline standard-baseline gaps. PFC is reported separately as a same-prefix evidence-redundancy audit. Multi-sample and process-supervised methods are contextual comparisons outside the online single-trace setting.}
\label{tab:method-scope}
\end{table}

\subsection{Evidence Regimes for Online Reliability Tracking}

\begin{table*}[t]
\centering
\scriptsize
\setlength{\tabcolsep}{3pt}
\resizebox{\linewidth}{!}{%
\begin{tabular}{@{}lrrrrrrr@{}}
\toprule
Setting & Err. & Obs. & Score & Brier $\Delta\downarrow$ & HC & Text/Self & LT/Act \\
\midrule
\multicolumn{8}{@{}l}{\textit{Main rows}} \\
MATH-500 Level 1 / 1.5B & 40.1 & 44.3 & -0.004 & -0.026 & +0.101 & +0.031 & \textbf{+0.136} \\
MATH-500 Level 2 / 1.5B & 45.8 & 43.5 & -0.015 & -0.041 & +0.041 & +0.020 & \textbf{+0.075} \\
MATH-500 Level 4 / 7B & 59.4 & 43.4 & -0.020 & -0.120 & +0.068 & \textbf{+0.104} & -- \\
MATH-500 Level 5 / 7B & 71.1 & 40.7 & +0.017 & -0.156 & +0.062 & \textbf{+0.110} & +0.067 \\
GSM8K / 7B & 28.3 & 30.5 & -0.007 & -0.010 & \textbf{+0.077} & +0.056 & +0.039 \\
RIMO-N / 14B & 46.6 & 21.2 & -0.043 & -0.163 & \textbf{+0.054} & +0.030 & -0.001 \\
RIMO-N Qwen3-14B & 43.7 & 18.5 & -0.065 & -0.133 & \textbf{+0.074} & +0.045 & -0.011 \\
\addlinespace[1pt]
\multicolumn{8}{@{}l}{\textit{Stress-test rows}} \\
AIME 2025 / 14B & 51.7 & 25.6 & -0.107 & -0.182 & -0.029 & -0.004 & -0.011 \\
Qwen3-14B / MATH-500 Level 5 & 47.2 & 17.2 & -0.024 & -0.123 & -0.023 & -0.021 & -- \\
\bottomrule
\end{tabular}}
\caption{Main split-seed results. Error rate (Err.) is trace-level final-answer error percentage, and Obs. is mean observations per trace. Score, HC, Text/Self, and LT/Act are compact headings for score-only filtering, hidden-cluster observations, text and self-verification markers, and latent-trajectory and activation features. Each AUROC entry is the mean gap versus the best standard prefix-safe baseline excluding the PFC audit. Appendix~\ref{sec:claim-evidence} reports post-PFC evidence-redundancy audit gaps. Brier $\Delta\downarrow$ is score-only \method{} Brier minus EMA. Bold marks the strongest positive structure-aware AUROC gap within each main row. Stress-test rows are left unbolded. Lower Brier deltas are better. Figure~\ref{fig:gaps} reports Brier improvement as the negative of this Brier delta.}
\label{tab:main}
\end{table*}

Table~\ref{tab:main} reveals a regime structure. Online belief tracking improves rank when prefix evidence is structure-aware and not already captured by the standard score/length baselines. The PFC audit then tests a stricter notion of same-prefix redundancy. Score-only filtering often improves probability quality but is rarely the best rank method. The most direct hard-regime row is MATH-500 Level 5 with DeepSeek-Qwen-7B, where text markers retain a $+0.069$ AUROC gap after the same-prefix PFC audit. RIMO-N adds cross-model evidence: self-verification is the more direct PFC-audited signal, while hidden-cluster observations are positive in the pre-PFC split-seed view and become mechanism evidence after PFC. AIME and Qwen-family MATH rows show where learned-prefix and length-aware baselines absorb rank signal.

\begin{figure}[t]
 \centering
 \includegraphics[width=0.94\columnwidth]{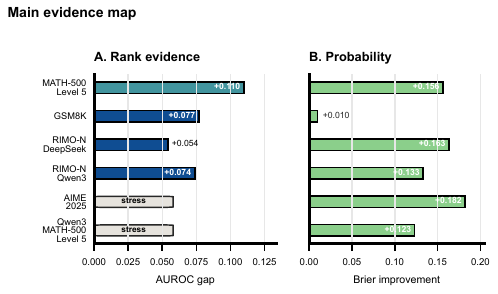}
 \caption{Main evidence map. Panel A reports only positive structure-aware AUROC gains over the standard prefix-safe baseline set, excluding the PFC audit. Rows without rank gain are marked as stress tests and omitted from the positive-gain panel. This compact positive-gain map is paired with the signed-gap view in Appendix~\ref{sec:appendix-diagnostics}. Stress rows remain visible in the signed evidence. Panel B reports score-only Brier improvement versus EMA, the negative of the Brier delta in Table~\ref{tab:main}. Larger positive values mean better probability quality.}
 \label{fig:gaps}
\end{figure}

Figure~\ref{fig:gaps} summarizes this pattern. Rank gains depend on the interaction between observation family, temporal integration, and prefix-safe baselines. The main figure reports positive structure-aware rows and marks AIME/Qwen-family rows as stress-test cases. Figure~\ref{fig:signed-auroc-boundary} in Appendix~\ref{sec:appendix-diagnostics} reports the full signed AUROC gaps.

\begin{figure}[t]
 \centering
 \includegraphics[width=0.94\columnwidth]{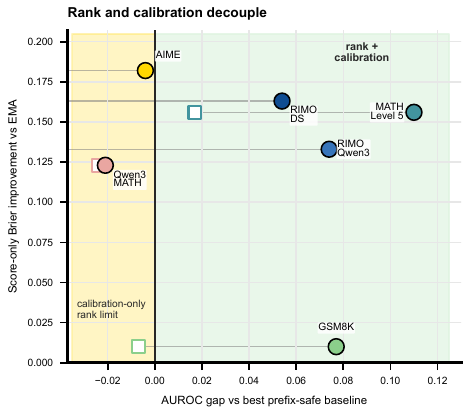}
 \caption{Rank and calibration are separate axes. Circles pair the strongest structure-aware AUROC gap against the standard prefix-safe baseline set with the score-only \method{} Brier improvement versus EMA. Hollow squares mark the corresponding score-only rank gap. MATH-500, GSM8K, and RIMO rows occupy the upper-right evidence region, while AIME and Qwen-family MATH show the stress-test pattern: probability quality can improve even when standard prefix-safe baselines dominate rank.}
 \label{fig:rank-calibration}
\end{figure}

Figure~\ref{fig:rank-calibration} separates the two axes behind this result. The horizontal axis measures whether an observation family improves ranking over the standard prefix-safe baseline set, and the vertical axis measures score-only Brier improvement versus EMA. The main rows occupy the rank-and-calibration quadrant, while AIME and Qwen-family MATH show that Brier improvements can coexist with negative rank gaps. The evaluation reports observation family, calibration, and baseline strength together, avoiding a single AUROC leaderboard.

\begin{figure}[t]
 \centering
 \includegraphics[width=0.94\columnwidth]{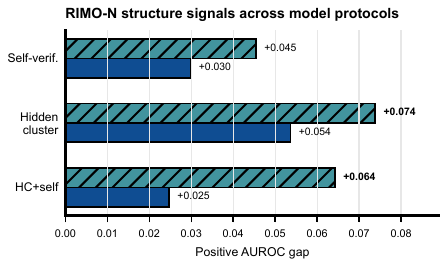}
 \caption{RIMO-N cross-model observation evidence. Blue bars are the DeepSeek-Qwen-14B row, and hatched teal bars are the Qwen3-14B protocol-aligned row. Bars report the pre-PFC split-seed view, where hidden-cluster observations, self-verification markers, and their joint variant are positive in both settings. The appendix PFC audit keeps self-verification as the more direct cross-model signal and treats hidden-cluster gains as mechanism evidence.}
 \label{fig:rimo-mechanism}
\end{figure}

Figure~\ref{fig:rimo-mechanism} isolates the RIMO-N cross-model slice. The bars show the pre-PFC split-seed view, where hidden-cluster observations and self-verification markers are positive in both 14B settings. The PFC audit then makes self-verification the more direct cross-model signal and treats hidden-cluster gains as mechanism evidence.

The PFC audit asks whether \method{} observations contain rank information not already absorbed by flexible score, length, and concept-code summaries. Positive post-PFC rows provide stronger evidence of non-redundant prefix signal, while rows absorbed by the classifier identify regimes where the evidence is predictive but not uniquely captured by Bayesian tracking. MATH-500 Level 5 text markers retain the largest post-PFC positive gap. For RIMO-N, split-seed best-variant self-verification is slightly positive on both 14B settings, while paired fixed-hybrid comparisons against PFC are negative or near-tie. Appendix~\ref{sec:claim-evidence} reports the full rows.

\subsection{Probability Quality, Calibration, and Utility}

Brier-score comparisons \citep{brier1950verification} are more favorable to score-only filtering than AUROC comparisons. On MATH-500 Level 5, score-only \method{} improves Brier versus EMA by $0.156$ while structure-aware variants provide rank gains. On AIME 2025, Brier improves even when AUROC remains negative. This separation is metric-level, not only empirical: AUROC depends on the ordering of scores across positive and negative traces, while Brier depends on the numerical probability assigned to the realized outcome. Appendix~\ref{sec:metric-algebra} gives the algebraic argument, and Brier versus calibrated last-step remains a stricter limitation for several rows.

Calibration-mode ablations help bound the all-prefix weak-supervision assumption. EM calibration improves rank on MATH-500 Levels 4/5 but does not fully resolve calibrated-last Brier limitations. Final-step calibration is weaker in the main rows. Table~\ref{tab:paired-calibration} separately reports paired deltas for the main and stress-test rows. This supports reporting all-step calibration as the default empirical setting, while making clear that it is weak supervision for eventual success, not true step labels.

\begin{table*}[t]
\centering
\scriptsize
\setlength{\tabcolsep}{3pt}
\renewcommand{\arraystretch}{1.08}
\begin{tabular}{@{}>{\raggedright\arraybackslash}p{0.25\linewidth}>{\raggedright\arraybackslash}p{0.09\linewidth}>{\raggedright\arraybackslash}p{0.13\linewidth}>{\raggedright\arraybackslash}p{0.24\linewidth}>{\raggedright\arraybackslash}p{0.19\linewidth}@{}}
\toprule
Setting & Metric & Signal & Reference & Delta [95\% CI] \\
\midrule
\multicolumn{5}{@{}l}{\textit{Intervals favoring \method{}}} \\
MATH-500 Level 1 / 1.5B & Brier & score & calibrated last-step & $\mathbf{-0.074}$ [$-0.131,-0.017$] \\
MATH-500 Level 5 / 7B & AUROC & hybrid & EMA & $\mathbf{+0.160}$ [$+0.022,+0.326$] \\
MATH-500 Level 5 / 7B & AUROC & hybrid & temporal baseline & $\mathbf{+0.261}$ [$+0.055,+0.470$] \\
RIMO-N DeepSeek-Qwen-14B & Brier & score & EMA & $\mathbf{-0.186}$ [$-0.248,-0.121$] \\
RIMO-N Qwen3-14B & Brier & score & EMA & $\mathbf{-0.159}$ [$-0.221,-0.093$] \\
Qwen3-14B / MATH-500 Level 5 & Brier & score & calibrated last-step & $\mathbf{-0.061}$ [$-0.123,-0.002$] \\
\addlinespace[1pt]
\multicolumn{5}{@{}l}{\textit{Stress-test and near-tie intervals}} \\
GSM8K / 7B & AUROC & hybrid & EMA & $+0.111$ [$-0.069,+0.269$] \\
RIMO-N DeepSeek-Qwen-14B & AUROC & hybrid & temporal baseline & $-0.221$ [$-0.353,-0.089$] \\
RIMO-N Qwen3-14B & AUROC & hybrid & EMA & $+0.002$ [$-0.119,+0.128$] \\
AIME 2025 / 14B & AUROC & hybrid & temporal baseline & $+0.000$ [$-0.281,+0.364$] \\
Qwen2.5 / MATH-500 Level 5 & AUROC & hybrid & temporal baseline & $-0.190$ [$-0.303,-0.095$] \\
\bottomrule
\end{tabular}
\caption{Compact paired-delta summary for the main and stress-test rows. Positive AUROC deltas and negative Brier deltas favor \method{} under the listed signal. Bold marks favorable paired deltas whose 95\% interval does not cross zero. Rows are computed from paired question-cluster bootstrap resamples and summarize paired evidence for the main-text claims and their limits.}
\label{tab:paired-calibration}
\end{table*}

Utility exports turn beliefs into early bad-trace detection thresholds chosen on calibration traces. Text and self-verification variants improve high-recall savings on MATH-500 Level 5. On RIMO-N, they save roughly $0.48$--$0.55$ of incorrect-trace compute at $0.45$--$0.51$ false-positive rates. We treat these results as decision-facing triage tradeoffs: thresholds are selected on calibration traces, operating points quantify early incorrect-trace detection, compute savings, and false positives, and the curves do not claim formal risk-control guarantees. Appendix Figure~\ref{fig:utility} summarizes the thresholded curves \citep{geifman2019selectivenet,schuster2022confident,chen2023frugalgpt}.

Appendix~\ref{sec:reliability-diagnostics} reports reliability diagrams and ECE summaries for the headline and stress-test rows. These curves support the calibration/ranking decoupling finding but also show a limitation: the identity state readout is not uniformly the best-calibrated probability estimate, especially on hard MATH and RIMO rows.

\subsection{Online and Offline Behavior}

Fixed-fraction prefix curves test whether the online belief is useful before the final prefix. The fixed-fraction exports evaluate 5\%, 25\%, 50\%, 75\%, and 100\% prefix fractions for MATH-500 Level 5, GSM8K, and both RIMO-N headline rows. Figure~\ref{fig:prefix-trajectory-ribbon} reports the full curves with question-cluster bootstrap intervals.

Fixed-fraction curves show two regimes. On MATH-500 Level 5, online hybrid AUROC grows from $0.500$ at 5\% and $0.586$ at 25\% to $0.733$ at 75\% and $0.786$ at the final prefix, while the raw prefix score is non-monotonic and falls to $0.503$ at the final prefix. GSM8K has weaker early evidence but reaches $0.662$ and $0.646$ at the 75\% and 100\% prefixes, with late-prefix Brier lower than the raw score. RIMO-N score/hybrid curves stay weak or declining even though hidden-cluster and self-verification final rows are positive. The contrast locates the failure mode: RIMO rank gain comes from structure-aware observations, not scalar score trajectories alone.

Offline smoothing and Viterbi are shown only as diagnostics because they can use future observations. Their gap from online filtering measures information still hidden in later prefixes. We use this only as an offline upper-bound diagnostic.

\subsection{Prefix-Value Validation Without Manual Process Labels}

Final-answer supervision cannot verify individual mathematical steps, so we add rollout validation as a prefix-value stress test inspired by sampled-continuation process supervision \citep{uesato2022solving,wang2024mathshepherd,luo2024improve}. For a held-out prefix $x_{1:t}$, we sample $K$ continuations from the same model and prompting protocol, score the continuation final answers, and estimate
\begin{equation}
 \hat v_t = \frac{1}{K}\sum_{k=1}^{K} r_{t,k},\qquad r_{t,k}\in\{0,1\}.
\end{equation}
Here $r_{t,k}=1$ iff continuation $k$ answers correctly, and $r_{t,k}=0$ otherwise. This estimates empirical continuation success under the chosen rollout policy and sampling protocol, not ground-truth mathematical recoverability. A binomial standard error is $\mathrm{SE}(\hat v_t)\approx\sqrt{\hat v_t(1-\hat v_t)/K}$. The diagnostic joins source prefix scores and online \method{} beliefs to rollout rows and does not identify the first wrong step.

The completed RIMO-N DeepSeek, RIMO-N Qwen3, and MATH-500 Level 5 belief joins are summarized in Table~\ref{tab:rollout-validation} and Figure~\ref{fig:rollout-belief-validation}: \method{} belief improves Brier score against empirical continuation success on all three held-out prefix sets, while rank correlations remain setting-dependent.

\section{Discussion}

\method{} provides a calibration-aware online inference framework for eventual-success reliability. The default identity readout is a reliability score. Calibrated target-probability readouts require an outcome readout or post-hoc calibration map. The value of the tracker depends on whether prefix-safe observations expose evidence not already absorbed by strong same-prefix baselines. The experiments reveal an evidence regime: structure-aware observations improve ranking when they add non-redundant prefix evidence, while score-only filtering more often improves probability quality. PFC audits evidence redundancy under flexible prefix classifiers, and rollout joins test whether belief trajectories transfer to empirical continuation success. The completed joins lower Brier on MATH-500 and RIMO-N, but rank association remains setting-dependent.

\section{Conclusion}

We frame LLM reasoning reliability as prefix-conditioned eventual-success tracking. \method{} separates observation design from temporal integration and shows that probability quality and rank can diverge. Structure-aware observations help most on MATH-500 and RIMO-N, while AIME and Qwen rows show where prefix-safe baselines absorb rank signal. For online monitoring, reasoning reliability should not be evaluated as a single static verifier score. Online monitors must distinguish probability-quality evidence from rank evidence and audit whether apparent gains survive strong prefix-safe baselines. SBBT makes these distinctions measurable under prefix safety.

\section*{Limitations}

The main limits follow from supervision and artifacts. A risk-focused manual final-answer audit found no label changes for the reported rows, but the supervision signal is still final-answer correctness, not human step-level/process annotation. Process-error claims need process-labeled data. Some AIME/Qwen rows show that competitive prefix-safe baselines can absorb available rank signal. The main deployment risk is misuse of these scores as standalone safety gates or pruning policies. Utility results are operating-point diagnostics without risk-control guarantees \citep{angelopoulos2024conformalriskcontrol}. Exact reproduction depends on generated trace artifacts, model/tokenizer versions, and answer-checking policy.

\bibliography{custom}

\appendix

\onecolumn
\raggedbottom
\section{Mathematical Details}
\label{sec:math-details}

This appendix records the probabilistic assumptions behind \method{} and the algebra used by the implementation.

\subsection{Notation Summary}

Table~\ref{tab:notation} lists the mathematical symbols used in the paper and appendix. Symbols that appear only as metric names, such as AUROC, Brier, and ECE, are defined where the corresponding metric is introduced.

\begin{table}[!ht]
\centering
\small
\setlength{\tabcolsep}{4pt}
\renewcommand{\arraystretch}{1.08}
\begin{tabular}{p{0.18\linewidth}p{0.76\linewidth}}
\toprule
Symbol & Meaning \\
\midrule
$q$ & Original problem or question. Splits are grouped by $q$. \\
$i$ & Trace index in held-out metric definitions. \\
$\tau$ & Calibration-trace index used in EM expected-count updates. \\
$t,T$ & Prefix index and number of prefixes in a trace. \\
$x_{1:t}$ & Generated text prefix available through prefix $t$. \\
$\phi$ & Prefix-safe observation function. \\
$o_t,o_{1:t},o_{1:T}$ & Observation at prefix $t$, observations through prefix $t$, and the full observation sequence. \\
$y$ & Final-answer correctness label; $y=1$ means the completed trace answers correctly. \\
$b_t$ & Online target probability $\Prob(y{=}1\,\vert\,o_{1:t})$. \\
$\Prob_{\theta}$, $\theta$ & Fitted two-state monitoring model and its parameters. \\
$z_t$ & Latent reliability state at prefix $t$. \\
$s,s',s^\star$ & Generic latent-state values in $\{\stateoff,\stateon\}$; $s^\star$ is a dummy summation state. \\
$\stateon,\stateoff$ & High-reliability and low-reliability latent states under the fitted monitoring model. \\
$\pi_t(s),\pi_0(s)$ & Filtered posterior $\Prob_{\theta}(z_t=s\mid o_{1:t})$ and initial state distribution. \\
$\tilde{\pi}_t(s)$ & Unnormalized filtering update used in the main-method equation. \\
$\hat b_t^{\mathrm{id}}$ & Identity state readout $\pi_t(\stateon)$. \\
$\rho_s$ & Outcome-readout probability $\Prob_{\theta}(y=1\mid z_t=s)$ or its calibration-split estimate. \\
$\eta_t(s)$ & Conditional outcome probability $\Prob_{\theta}(y=1\mid z_t=s,o_{1:t})$. \\
$b_t^{\theta}$ & Target probability under the fitted monitoring model. \\
$\hat b_t^{\rho}$ & Calibrated two-state outcome readout $\sum_s\rho_s\pi_t(s)$. \\
$g$ & Post-hoc calibration map from $\pi_t(\stateon)$ to a target-probability readout. \\
$\Thetapre$ & All parameters, observation extractors, readout maps, thresholds, baselines, and audit classifiers fitted before held-out test evaluation. \\
$\calF_t$ & Prefix information set $\sigma(\Thetapre,o_1,\ldots,o_t)$ for a test trace. \\
$A_{s's},A_t$ & Stationary transition entry $\Prob(z_t=s\mid z_{t-1}=s')$ and optional time-dependent transition matrix. \\
$p_{\mathrm{error}},p_{\mathrm{recover}}$ & Two-state transition probabilities for $\stateon\!\to\!\stateoff$ and $\stateoff\!\to\!\stateon$. \\
$\ell_s(o_t)$ & Emission likelihood value: a probability mass for discrete observations or a density value for continuous scores. \\
$\mu_t(s)$ & Predictive belief $\Prob_{\theta}(z_t=s\mid o_{1:t-1})$ before observing $o_t$. \\
$u_t(s)$ & Unnormalized posterior mass $\ell_s(o_t)\mu_t(s)$. \\
$\mathcal{Z}_t$ & Filtering normalizer $\sum_{s^\star}u_t(s^\star)$. \\
$a_t,d_t,p_s(a_t)$ & Continuous score observation, discrete concept-code observation, and state-conditioned score density. \\
$\Bin(\cdot)$ & Discretization map for continuous scores. \\
$\alpha_t(s),\beta_t(s)$ & Forward and backward messages in the EM calibration derivation. \\
$\gamma_t(s)$ & Smoothed state responsibility $\Prob_{\theta}(z_t=s\mid o_{1:T})$. \\
$\xi_t(s,s')$ & Pairwise transition responsibility used only if transitions are re-estimated. \\
$k,j,N_{s,k},\lambda$ & Score-bin index, score-bin summation index, expected smoothed count, and additive smoothing constant. \\
$\tilde z_t$ & Surrogate all-prefix label equal to final trace label $y$ during weak-supervision calibration. \\
$n,\hat b_i,y_i$ & Number of evaluated traces, predicted probability for trace $i$, and final label for trace $i$. \\
$m,r,\Delta^{\Std}_{m,r},R$ & Method index, split seed, standard-baseline AUROC gap, and number of valid split seeds. \\
$\mathcal{B}^{\Std}_r,\mathcal{B}^{\Audit}_r,\PFC_r$ & Standard prefix-safe baseline set, audit baseline set after adding PFC, and the prefix-feature classifier for split $r$. \\
$\mathbf{1}\{\cdot\}$ & Indicator function. \\
\bottomrule
\end{tabular}
\caption{Notation used by the SBBT derivations and metrics.}
\label{tab:notation}
\end{table}

\subsection{Model and Target}

Let $y \in \{0,1\}$ denote final-answer correctness for a generated trace, and let $o_{1:T}$ denote the prefix observations extracted from that trace. The online target at prefix $t$ is
\begin{equation}
 b_t = \Prob(y{=}1\,\vert\,o_{1:t}).
\end{equation}
\method{} first estimates a posterior over a two-state latent reliability variable,
\begin{equation}
 \pi_t(s)=\Prob_{\theta}(z_t=s \mid o_{1:t}),
 \qquad s\in\{\stateoff,\stateon\}.
\end{equation}
Here $\Prob_{\theta}$ denotes the fitted two-state monitoring model. The online score used in the experiments is the identity state readout
\begin{equation}
 \hat b_t^{\mathrm{id}}=\pi_t(\stateon).
\end{equation}
This is a model-based reliability score, not a theorem that $z_t$ is literally the final label $y$. The high-reliability state $\stateon$ means that the prefix observation is more likely under traces that eventually succeed, not that every mathematical step up to $t$ is externally verified.

\subsection{Readout Versus Target Probability}

The mathematically general way to connect the latent posterior to the target $b_t$ is to add an outcome readout. Let
\begin{equation}
 \rho_s=\Prob_{\theta}(y=1\mid z_t=s),
 \qquad s\in\{\stateoff,\stateon\},
\end{equation}
or let $\rho_s$ denote a calibration-split estimate of that conditional outcome probability. We write $\rho_s$ as a time-homogeneous readout for simplicity; a more general readout could use $\rho_{t,s}$ when the outcome-state relation changes with prefix position. Define
\begin{equation}
 \eta_t(s)=\Prob_{\theta}(y=1\mid z_t=s,o_{1:t}).
\end{equation}
Then the law of total probability gives
\begin{align}
 b_t^{\theta}
 &= \Prob_{\theta}(y=1\mid o_{1:t}) \\
 &= \sum_s \Prob_{\theta}(y=1,z_t=s\mid o_{1:t}) \\
 &= \sum_s \eta_t(s)\,\pi_t(s).
\end{align}
If the extended monitoring model further assumes $y\perp o_{1:t}\mid z_t$, then $\eta_t(s)=\rho_s$, which gives
\begin{equation}
 b_t^{\theta}=\sum_s \rho_s\,\pi_t(s).
\end{equation}
Under this additional assumption, a calibrated two-state probability readout is
\begin{equation}
 \hat b_t^{\rho}=\rho_{\stateon}\pi_t(\stateon)+\rho_{\stateoff}\pi_t(\stateoff).
\end{equation}
The identity readout $\hat b_t^{\mathrm{id}}=\pi_t(\stateon)$ is the special case $\rho_{\stateon}=1$ and $\rho_{\stateoff}=0$. That special case is exact only under deterministic state-label identification. We use it as the default reliability score and evaluate it empirically with Brier and calibration diagnostics.

A still more flexible target-probability readout can be written as
\begin{equation}
 \hat b_t(g) = g(\pi_t(\stateon)),
\end{equation}
where $g:[0,1]\rightarrow[0,1]$ is a monotone or unconstrained post-hoc calibration map fitted only on calibration prefixes \citep{guo2017calibration}. In this sense, \method{} is a calibration-aware online inference framework: the default identity readout is evaluated as a reliability score, while calibrated target-probability readouts require an additional outcome readout or post-hoc calibration map. The experiments support calibration-aware online reliability estimation through measured probability quality. They do not assert that the hidden state itself is a ground-truth correctness label.

\subsection{Forward Filtering Derivation}

Assume a first-order transition model and a conditionally Markov observation model:
\begin{align}
 \Prob(z_t=s \mid z_{1:t-1}) &= \Prob(z_t=s \mid z_{t-1}), \\
 \Prob(o_t \mid z_{1:t}, o_{1:t-1}) &= \Prob(o_t \mid z_t).
\end{align}
Because the observations are engineered from cumulative reasoning prefixes, the second line is a modeling approximation over the observation sequence, not a claim that the generated text itself is conditionally independent. Let $A_{s's}=\Prob(z_t=s \mid z_{t-1}=s')$, let $\ell_s(o_t)$ denote the state-conditioned emission likelihood value, and let $\pi_t(s)=\Prob_{\theta}(z_t=s \mid o_{1:t})$. For discrete observations, $\ell_s(o_t)$ is a probability mass; for continuous scores, it is a density or binned likelihood value. Let $\mu_t(s)=\Prob_{\theta}(z_t=s\mid o_{1:t-1})$ denote the predictive belief before seeing $o_t$. For $t=1$, $\mu_1(s)$ is the pre-fit initial distribution $\pi_0(s)$. For $t>1$,
\begin{equation}
 \mu_t(s)=\sum_{s'} A_{s's}\,\pi_{t-1}(s').
\end{equation}
By Bayes' rule and the conditional observation model, the unnormalized posterior mass after seeing $o_t$ is
\begin{equation}
 u_t(s)=\ell_s(o_t)\mu_t(s).
\end{equation}
Let
\begin{equation}
 \mathcal{Z}_t=\sum_{s^\star}u_t(s^\star)
 =\sum_{s^\star}\ell_{s^\star}(o_t)\mu_t(s^\star).
\end{equation}
When $\mathcal{Z}_t>0$, the normalized filtering belief is
\begin{equation}
 \pi_t(s)=\frac{u_t(s)}{\mathcal{Z}_t}.
\end{equation}
Substituting the definition of $\mu_t$ gives the expanded update in the main text. This is the standard HMM forward filtering recursion \citep{rabiner1989tutorial,sarkka2013bayesian}. In the two-state implementation, $A_{\stateon,\stateoff}=p_{\mathrm{error}}$, $A_{\stateoff,\stateon}=p_{\mathrm{recover}}$, $A_{\stateon,\stateon}=1-p_{\mathrm{error}}$, and $A_{\stateoff,\stateoff}=1-p_{\mathrm{recover}}$. A time-dependent transition ablation replaces $A$ by $A_t$ in the same equations.

\paragraph{Normalization.}
If the emission likelihoods and transition probabilities are nonnegative and $\mathcal{Z}_t>0$, then $\sum_s \pi_t(s)=1$ and $\pi_t(s)\ge 0$. This follows immediately because each posterior mass is divided by $\mathcal{Z}_t=\sum_s u_t(s)$. With additive likelihood smoothing and a normalized $\pi_0$, $\mathcal{Z}_t$ is positive whenever at least one state has positive predictive mass and positive smoothed emission likelihood. This normalization is internal to the monitoring model. It does not by itself prove perfect calibration for $y$, so probability quality is evaluated with Brier/ECE-style diagnostics and calibrated baselines. If a numerical normalizer is zero in implementation, the filter falls back to the predictive belief without adding new evidence.

\subsection{Log-Odds Evidence Form}
\label{sec:log-odds-evidence}

\paragraph{Proposition.}
Assume the two-state filtering update above and suppose
\[
\mu_t(\stateon)>0,\quad
\mu_t(\stateoff)>0,\quad
\ell_{\stateon}(o_t)>0,\quad
\ell_{\stateoff}(o_t)>0.
\]
Then the posterior log-odds decompose into predictive log-odds plus observation evidence:
\begin{equation}
\log \frac{\pi_t(\stateon)}{\pi_t(\stateoff)}
=
\log \frac{\mu_t(\stateon)}{\mu_t(\stateoff)}
+
\log \frac{\ell_{\stateon}(o_t)}{\ell_{\stateoff}(o_t)}.
\end{equation}
If the transition is the identity transition, so that $\mu_t(s)=\pi_{t-1}(s)$ for both states, then
\begin{equation}
\log \frac{\pi_t(\stateon)}{\pi_t(\stateoff)}
=
\log \frac{\pi_0(\stateon)}{\pi_0(\stateoff)}
+
\sum_{k=1}^{t}
\log \frac{\ell_{\stateon}(o_k)}{\ell_{\stateoff}(o_k)}.
\end{equation}

\paragraph{Proof.}
The common normalizer cancels in the odds ratio. Since $u_t(s)=\ell_s(o_t)\mu_t(s)$ and $\pi_t(s)=u_t(s)/\mathcal{Z}_t$,
\begin{align}
\frac{\pi_t(\stateon)}{\pi_t(\stateoff)}
&=
\frac{u_t(\stateon)}{u_t(\stateoff)}\\
&=
\frac{\ell_{\stateon}(o_t)\mu_t(\stateon)}
 {\ell_{\stateoff}(o_t)\mu_t(\stateoff)}\\
&=
\frac{\mu_t(\stateon)}{\mu_t(\stateoff)}
\cdot
\frac{\ell_{\stateon}(o_t)}{\ell_{\stateoff}(o_t)}.
\end{align}
The positivity assumptions make the logarithm valid, giving the one-step identity. If the transition is the identity transition, then $\mu_t(s)=\pi_{t-1}(s)$ for both states, and the same identity becomes
\begin{equation}
\log \frac{\pi_t(\stateon)}{\pi_t(\stateoff)}
=
\log \frac{\pi_{t-1}(\stateon)}{\pi_{t-1}(\stateoff)}
+
\log \frac{\ell_{\stateon}(o_t)}{\ell_{\stateoff}(o_t)}.
\end{equation}
Iterating this recurrence from $1$ through $t$ yields
\begin{equation}
\log \frac{\pi_t(\stateon)}{\pi_t(\stateoff)}
=
\log \frac{\pi_0(\stateon)}{\pi_0(\stateoff)}
+
\sum_{k=1}^{t}
\log \frac{\ell_{\stateon}(o_k)}{\ell_{\stateoff}(o_k)}.
\end{equation}
This proves the accumulated-evidence form. \hfill$\square$

\paragraph{Consequence.}
The quantity
\begin{equation}
\lambda_t(o_t)=
\log \frac{\ell_{\stateon}(o_t)}{\ell_{\stateoff}(o_t)}
\end{equation}
is the calibrated log-likelihood-ratio evidence contributed by prefix $t$. Positive values support the high-reliability state, negative values support the low-reliability state, and values near zero are weak evidence under the fitted monitoring model. With non-identity transitions, the same evidence term is added after a transition-prediction step that can damp or shift prior odds before incorporating the new observation.

\subsection{Hybrid and Joint Observation Likelihoods}

For score-only observations, $\ell_s(o_t)$ is a calibrated probability mass over a score bin or a density value over a score range. For concept-only observations, it is a smoothed categorical likelihood over concept codes. The hybrid variant uses
\begin{equation}
 \ell_s(o_t) \approx p_s(a_t)\Prob(d_t \mid z_t=s),
\end{equation}
where $a_t$ is a continuous score observation, $p_s(a_t)$ is the state-conditioned score density, and $d_t$ is a discrete concept code. If the implementation uses binned scores, this corresponds to $\Prob(\Bin(a_t)\mid z_t=s)\Prob(d_t\mid z_t=s)$. This is a conditional-independence approximation, not a theorem about the data-generating process. The joint variant instead estimates
\begin{equation}
 \ell_s(o_t)=\Prob(\Bin(a_t), d_t \mid z_t=s)
\end{equation}
directly, trading a weaker independence assumption for a larger discrete likelihood table.

\subsection{EM Emission Calibration}

The EM calibration variant follows the standard Baum-Welch pattern for the score-emission histogram \citep{baum1970maximization,dempster1977maximum,rabiner1989tutorial}. For a calibration trace, define forward and backward messages under the current parameters:
\begin{align}
 \alpha_1(s) &= \pi_0(s)\ell_s(o_1), && t=1, \\
 \alpha_t(s) &= \ell_s(o_t)\sum_{s'}\alpha_{t-1}(s')A_{s's}, && 1<t\le T, \\
 \beta_T(s) &= 1, && t=T, \\
 \beta_t(s) &= \sum_{s'} A_{ss'}\ell_{s'}(o_{t+1})\beta_{t+1}(s'), && 1\le t<T.
\end{align}
The marginal likelihood of the observation sequence is
\begin{equation}
 \Prob_{\theta}(o_{1:T})=\sum_s \alpha_T(s).
\end{equation}
The smoothed state responsibility is
\begin{equation}
 \gamma_t(s)
 = \Prob_{\theta}(z_t=s \mid o_{1:T})
 = \frac{\alpha_t(s)\beta_t(s)}
 {\sum_{s^\star} \alpha_t(s^\star)\beta_t(s^\star)}.
\end{equation}
For score bin $k$, trace index $\tau$, and score observation $a_{\tau,t}$, the M-step updates smoothed histogram counts as
\begin{align}
 N_{s,k}
 &= \lambda + \sum_{\tau}\sum_t
 \gamma_{\tau,t}(s)\,\mathbf{1}\{\Bin(a_{\tau,t})=k\}, \\
 \ell_s(k)
 &=\frac{N_{s,k}}{\sum_j N_{s,j}}.
\end{align}
Here $\lambda>0$ is an additive smoothing constant. The implementation keeps transition parameters fixed in this calibration variant and re-estimates score emissions from these expected counts, initialized from the all-prefix outcome labels. If transitions were also re-estimated, the standard update would use pairwise responsibilities $\xi_t(s,s')=\Prob_{\theta}(z_t=s,z_{t+1}=s'\mid o_{1:T})$. We do not use that update for the evaluated EM calibration variant. As usual for EM, Baum-Welch is a local likelihood-fitting procedure and does not guarantee that the fitted states recover human step-correctness.

The two latent states are exchangeable under the likelihood, so EM fitting also needs an orientation step before reporting an \stateon{} readout. On calibration traces we compute
\begin{equation}
 \hat\rho_s
 =
 \frac{\sum_{\tau,t}\gamma_{\tau,t}(s)y_\tau}
 {\sum_{\tau,t}\gamma_{\tau,t}(s)},
 \qquad
 \stateon=\arg\max_s \hat\rho_s.
\end{equation}
After EM emission fitting, we orient the two states by assigning \stateon{} to the state with larger expected final-success rate $\hat\rho_s$ on calibration traces. This removes label-switching ambiguity for the reported identity readout without turning the state into an oracle step label.

\subsection{Prefix-Safety as Adaptedness}

\paragraph{Pre-fit parameter set.}
Let $\Thetapre$ denote all quantities fitted before held-out test evaluation. This includes the initial belief $\pi_0$, transition parameters, emission likelihoods, score orientations, cluster assignments, probe parameters, calibration maps, readout maps, utility thresholds, learned-prefix baselines, and prefix-feature classifiers. By protocol, $\Thetapre$ is fitted only on train/calibration questions and is fixed during test evaluation.

For a test trace, define the prefix information set
\begin{equation}
\calF_t=\sigma(\Thetapre,o_1,\ldots,o_t).
\end{equation}
Equivalently, because each observation is computed from the generated prefix, $\calF_t$ is the information generated by pre-fit quantities and all prefix-safe observations available through time $t$. The information sets are nested:
\begin{equation}
\calF_0\subseteq \calF_1\subseteq \cdots \subseteq \calF_T.
\end{equation}

\paragraph{Proposition.}
Suppose every test-time observation $o_t$ is computed only from the generated prefix $x_{1:t}$ and metadata available no later than prefix $t$. Suppose all fitted quantities are contained in $\Thetapre$ and are fixed before test evaluation. Then, for each prefix $t$, the filtered belief $\pi_t(\stateon)$ is $\calF_t$-measurable. If $g$ is a pre-fit readout map, then $g(\pi_t(\stateon))$ is also $\calF_t$-measurable.

\paragraph{Proof.}
The claim follows by induction. At $t=1$, $\pi_0$, transition parameters, emission likelihood tables or densities, and any readout map are contained in $\Thetapre$. The observation $o_1$ is contained in $\calF_1$, so $\ell_s(o_1)$, $u_1(s)=\ell_s(o_1)\mu_1(s)$, the normalizer $\mathcal{Z}_1=\sum_{s^\star}u_1(s^\star)$, and $\pi_1(s)=u_1(s)/\mathcal{Z}_1$ are all $\calF_1$-measurable whenever $\mathcal{Z}_1>0$.

For the induction step, assume $\pi_{t-1}(s)$ is $\calF_{t-1}$-measurable for both states. Since $\calF_{t-1}\subseteq\calF_t$, the previous belief is also $\calF_t$-measurable. With pre-fit transition entries $A_{s's}$, the predictive belief is
\begin{equation}
\mu_t(s)=\sum_{s'}A_{s's}\pi_{t-1}(s').
\end{equation}
The observation $o_t$ is included in $\calF_t$, and the emission function $\ell_s(\cdot)$ is pre-fit, so
\begin{equation}
u_t(s)=\ell_s(o_t)\mu_t(s)
\end{equation}
and
\begin{equation}
\mathcal{Z}_t=\sum_{s^\star}u_t(s^\star)
\end{equation}
are $\calF_t$-measurable. Finite sums, products, likelihood lookups, and normalizations preserve measurability, so when $\mathcal{Z}_t>0$,
\begin{equation}
\pi_t(s)=\frac{u_t(s)}{\mathcal{Z}_t}
\end{equation}
is $\calF_t$-measurable. In particular, $\pi_t(\stateon)$ is $\calF_t$-measurable. If $g$ is pre-fit, then $g$ is fixed in $\Thetapre$ before test evaluation, and the composition $g(\pi_t(\stateon))$ is also $\calF_t$-measurable. \hfill$\square$

\paragraph{Online consequence and hidden features.}
The proposition says that test-time prediction remains adapted to the prefix filtration: the online score at prefix $t$ is a function only of pre-fit quantities and observations available through $t$. For hidden-state features extracted from completed generations, the adaptedness argument requires causal extraction. We use causal hidden states or pooled token spans corresponding to the prefix window; no future-token hidden vectors enter $o_t$. If a hidden feature uses future-token hidden vectors, then the premise $o_t\in\calF_t$ fails and the resulting score is not prefix-safe.

\subsection{Why Smoothing and Viterbi Are Offline Checks}

Forward-backward smoothing estimates
\begin{equation}
 \Prob(z_t=s \mid o_{1:T})
 = \frac{\alpha_t(s)\beta_t(s)}
 {\sum_{s^\star} \alpha_t(s^\star)\beta_t(s^\star)},
\end{equation}
where the backward message satisfies
\begin{equation}
 \beta_t(s)=\sum_{s'} A_{ss'}\ell_{s'}(o_{t+1})\beta_{t+1}(s').
\end{equation}
Because $\beta_t$ contains likelihood terms from $o_{t+1:T}$, smoothing uses future observations unavailable at prefix $t$. The Viterbi diagnostic used here similarly returns the most likely full state path under $o_{1:T}$ and can use observations after $t$ when deciding earlier states. A prefix-restricted Viterbi path could be computed online, but that is not the offline full-trace diagnostic reported here. Smoothing and full-trace Viterbi are useful diagnostics for whether future prefixes contain signal, but they are not online methods for early intervention or pruning.

\subsection{Weak Supervision From Final Labels}

The all-prefix calibration mode assigns the final trace label $y$ to each prefix observation during likelihood fitting. Let $\tilde{z}_t=y$ denote this surrogate prefix label. The fitted outcome-conditioned likelihood estimates
\begin{equation}
 \Prob(o_t \mid \tilde{z}_t=y),
\end{equation}
not an estimate of $\Prob(o_t \mid \text{current step is correct})$. This distinction matters: a trace can eventually answer correctly after a locally unproductive prefix, and a plausible prefix can later lead to a wrong answer. The transition dynamics then impose a temporal reliability model on top of these outcome-conditioned emissions. Under this supervision, all-prefix calibration is a weak-supervision scheme for eventual-success monitoring. Step-correctness or first-error claims require external process labels and separate validation.

\subsection{Question-Level Split Invariant}

Let $q$ index the original problem and let multiple traces be sampled for the same $q$. If traces from the same question appear in both fitting and test partitions, then observation distributions, answer formats, and difficulty cues can leak across partitions even when generated trace text differs. Grouping by $q$ ensures that calibration/test evaluation measures transfer to unseen questions instead of reusing question-specific structure. Splitting, calibration, clustering, probing, and threshold selection all use question-level partitions.

\subsection{Metric Definitions}

For a test split with $n$ traces, a probabilistic readout $\hat b_i$ and final label $y_i$, the Brier score is
\begin{equation}
 \Brier(\hat b,y)=\frac{1}{n}\sum_{i=1}^{n}(\hat b_i-y_i)^2.
\end{equation}
For method $m$ and split seed $r$, the main Table~\ref{tab:main} AUROC gap uses the standard prefix-safe baseline set $\mathcal{B}^{\Std}_r$:
\begin{equation}
 \Delta^{\Std}_{m,r}
 = \AUROC_{m,r}
 - \max_{b\in\mathcal{B}^{\Std}_r}\AUROC_{b,r}.
\end{equation}
The stricter PFC audit uses
\begin{equation}
 \mathcal{B}^{\Audit}_r
 = \mathcal{B}^{\Std}_r \cup \{\PFC_r\},
\end{equation}
and is reported separately in Appendix~\ref{sec:claim-evidence}. The Table~\ref{tab:main} headline gaps and the post-PFC audit gaps answer different questions: the former compares against standard online baselines, while the latter tests whether same-prefix score, length, and concept summaries absorb the rank signal.
AUROC is defined only for evaluation splits containing both final-success classes. Degenerate splits are not used for AUROC claims. Main tables report the mean standard-baseline gap $\frac{1}{R}\sum_r \Delta^{\Std}_{m,r}$ over the valid split seeds and the positive fraction $\frac{1}{R}\sum_r \mathbf{1}\{\Delta^{\Std}_{m,r}>0\}$. Brier deltas are always written as method minus baseline, so negative values indicate better probability quality.

\subsection{Metric Algebra for Ranking and Probability Quality}
\label{sec:metric-algebra}

\paragraph{Empirical AUROC.}
Let $P=\{i:y_i=1\}$ be the set of positive traces and $N=\{j:y_j=0\}$ be the set of negative traces. For a scalar score vector $r=(r_1,\ldots,r_n)$, define empirical AUROC as
\begin{equation}
\AUROC(r,y)
=
\frac{1}{|P||N|}
\sum_{i\in P}\sum_{j\in N}
\left[
\mathbf{1}\{r_i>r_j\}
+
\frac{1}{2}\mathbf{1}\{r_i=r_j\}
\right].
\end{equation}
This pairwise form makes explicit that AUROC is a rank statistic \citep{fawcett2006introduction,hand2001simple}.

\paragraph{Proposition.}
If $g:\mathbb{R}\rightarrow\mathbb{R}$ is strictly increasing, then
\begin{equation}
\AUROC(g(r),y)=\AUROC(r,y),
\end{equation}
where $g(r)=(g(r_1),\ldots,g(r_n))$.

\paragraph{Proof.}
Fix any positive trace $i\in P$ and any negative trace $j\in N$. A strictly increasing $g$ preserves both strict order and equality:
\[
r_i>r_j \Longleftrightarrow g(r_i)>g(r_j),
\qquad
r_i=r_j \Longleftrightarrow g(r_i)=g(r_j).
\]
The pairwise AUROC contribution is unchanged:
\begin{align}
&\mathbf{1}\{g(r_i)>g(r_j)\}
+
\frac{1}{2}\mathbf{1}\{g(r_i)=g(r_j)\}\\
&\qquad =
\mathbf{1}\{r_i>r_j\}
+
\frac{1}{2}\mathbf{1}\{r_i=r_j\}.
\end{align}
Summing this equality over all positive-negative pairs and dividing by $|P||N|$ gives $\AUROC(g(r),y)=\AUROC(r,y)$. \hfill$\square$

\paragraph{Brier score is not rank-invariant.}
For probability predictions $\hat b_i\in[0,1]$, the Brier score is
\begin{equation}
\Brier(\hat b,y)=\frac{1}{n}\sum_{i=1}^{n}(\hat b_i-y_i)^2.
\end{equation}
Unlike AUROC, Brier score depends on the numerical distance between the forecast and the realized binary label \citep{brier1950verification,gneiting2007strictly}.

For any strictly increasing map $g$, the proposition above fixes the AUROC value but leaves the Brier difference
\begin{equation}
\Brier(g(\hat b),y)-\Brier(\hat b,y)
=
\frac{1}{n}\sum_{i=1}^{n}
\left[(g(\hat b_i)-y_i)^2-(\hat b_i-y_i)^2\right]
\end{equation}
unconstrained by ranking. The sign depends on how $g$ moves the probability values relative to the realized labels: it can be positive, negative, or zero while the positive-negative ordering is unchanged. Brier improvement and AUROC improvement need not coincide. Strictly increasing calibration maps preserve AUROC; non-strict monotone maps can alter AUROC through tie creation. This metric behavior motivates reporting probability quality and ranking as separate axes.

\subsection{Evidence-Redundancy Audit Algebra}
\label{sec:pfc-audit-algebra}

For split seed $r$, let $\mathcal{B}^{\Std}_r$ be the standard prefix-safe baseline set used for the main headline gap. Let $\PFC_r$ denote the prefix-feature classifier fitted on calibration traces for the same split. Define
\begin{equation}
M^{\Std}_r=\max_{b\in \mathcal{B}^{\Std}_r}\AUROC_{b,r},
\end{equation}
and
\begin{equation}
M^{\Audit}_r
=
\max_{b\in \mathcal{B}^{\Std}_r\cup\{\PFC_r\}}\AUROC_{b,r}.
\end{equation}
For an online method or observation variant $m$, define the standard gap
\begin{equation}
\Delta^{\Std}_{m,r}
=
\AUROC_{m,r}-M^{\Std}_r,
\end{equation}
the direct PFC gap
\begin{equation}
\Delta^{\PFC}_{m,r}
=
\AUROC_{m,r}-\AUROC_{\PFC,r},
\end{equation}
and the audit gap
\begin{equation}
\Delta^{\Audit}_{m,r}
=
\AUROC_{m,r}-M^{\Audit}_r.
\end{equation}

\paragraph{Proposition.}
For every method $m$ and split seed $r$,
\begin{equation}
\Delta^{\Audit}_{m,r}
=
\min\left\{\Delta^{\Std}_{m,r},\Delta^{\PFC}_{m,r}\right\}.
\end{equation}
Consequently,
\begin{equation}
\Delta^{\Audit}_{m,r}>0
\quad\Longleftrightarrow\quad
\Delta^{\Std}_{m,r}>0
\ \text{and}\
\Delta^{\PFC}_{m,r}>0.
\end{equation}

\paragraph{Proof.}
The audit baseline set contains exactly the standard baseline set plus the PFC classifier, so
\begin{equation}
M^{\Audit}_r
=
\max\left\{
M^{\Std}_r,
\AUROC_{\PFC,r}
\right\}.
\end{equation}
Then
\begin{align}
\Delta^{\Audit}_{m,r}
&=
\AUROC_{m,r}-M^{\Audit}_r\\
&=
\AUROC_{m,r}
-
\max\left\{
M^{\Std}_r,
\AUROC_{\PFC,r}
\right\}.
\end{align}
We use the elementary identity
\[
a-\max\{b,c\}=\min\{a-b,a-c\}.
\]
which follows by considering whether $b\ge c$ or $c>b$. Applying this identity with $a=\AUROC_{m,r}$, $b=M^{\Std}_r$, and $c=\AUROC_{\PFC,r}$ gives
\begin{align}
\Delta^{\Audit}_{m,r}
&=
\min\left\{
\AUROC_{m,r}-M^{\Std}_r,\,
\AUROC_{m,r}-\AUROC_{\PFC,r}
\right\}\\
&=
\min\left\{\Delta^{\Std}_{m,r},\Delta^{\PFC}_{m,r}\right\}.
\end{align}
The minimum of two real numbers is positive exactly when both numbers are positive, which gives the stated equivalence. \hfill$\square$

This identity is per split seed. After averaging over split seeds, $\frac{1}{R}\sum_r \Delta^{\Audit}_{m,r}$ is not generally equal to the minimum of the two averaged gaps, $\min\{\frac{1}{R}\sum_r\Delta^{\Std}_{m,r},\frac{1}{R}\sum_r\Delta^{\PFC}_{m,r}\}$. The tables report the averaged audit gap directly instead of deriving it from averaged standard and PFC gaps.

\paragraph{Audit meaning.}
The standard gap asks whether an observation variant improves over the standard prefix-safe baseline set. The direct PFC gap asks whether it improves over a flexible same-prefix classifier using score, length, and concept-code summaries. The audit gap is positive only when both tests are passed. A positive post-PFC gap is evidence of nonredundant rank signal relative to the tested PFC feature class. A non-positive post-PFC gap does not mean the observation family is useless; it means that the available rank signal may already be absorbed by the same-prefix classifier or by the standard baselines.

\section{Claim/Evidence Alignment}
\label{sec:claim-evidence}

Table~\ref{tab:claim-evidence-map} matches each main claim to its evidence. Main evidence rows support the observation-family story, and stress-test rows delimit the scope.

\begin{table}[!ht]
\centering
\scriptsize
\setlength{\tabcolsep}{2pt}
\renewcommand{\arraystretch}{1.07}
\begin{tabular}{@{}>{\raggedright\arraybackslash}p{0.21\linewidth}>{\raggedright\arraybackslash}p{0.36\linewidth}>{\raggedright\arraybackslash}p{0.25\linewidth}>{\raggedright\arraybackslash}p{0.12\linewidth}@{}}
\toprule
Claim & Completed evidence & Remaining risk & Status \\
\midrule
The target is online eventual-success monitoring, not step verification. & Task definition, notation, weak-supervision appendix, and rollout-prefix validation all use final-answer success as the supervised target. & Process-error or first-error claims require external process labels and are not made here. & supported \\
\method{} is a calibration-aware tracker with an observation-agnostic interface. & The same filtering equations consume score, text and self-verification markers, hidden-cluster observations, token-pooling probes, and latent-trajectory observations. & Observation design explains most rank behavior beyond the shared two-state filter. & supported \\
Calibration and ranking decouple. & Table~\ref{tab:main}, Figure~\ref{fig:rank-calibration}, Table~\ref{tab:rollout-validation}, and Appendix~\ref{sec:reliability-diagnostics} show Brier/reliability behavior even when rank remains weak. & Identity-readout ECE remains setting-dependent, so reliability curves stay diagnostic, not headline evidence. & supported \\
Structure-aware observations drive the clearest rank gains. & MATH-500 Level 5 text markers, self-verification markers, and hidden-cluster rows remain positive in the split-seed PFC audit; RIMO self-verification remains slightly positive under the split-seed best-variant audit. & RIMO hidden-cluster rows are absorbed by PFC and treated as mechanism evidence. & supported \\
Stress-test rows are part of the result. & AIME and Qwen-family MATH rows are clean, non-degenerate cases where learned-prefix or length-aware baselines dominate AUROC. & These rows delimit the claim and remain in the appendix evidence. & supported \\
Reproducibility follows an explicit reproduction record. & Appendix~\ref{sec:reproducibility} lists the split protocol, artifact classes, checksum metadata, and known reproduction risks. & Run manifests and answer-audit summaries positive claims that depend on exact generated traces. & documented \\
\bottomrule
\end{tabular}
\caption{Claim/evidence map for reported claims.}
\label{tab:claim-evidence-map}
\end{table}

\paragraph{Marker-rule details.}
Table~\ref{tab:marker-rule-details} records the deterministic text-marker rules used by the text and self-verification observation families. These rules were fixed before evaluation and are not tuned on held-out rows. Marker extraction scans only the model-generated continuation contained in the current prefix, not the original problem prompt.

\begin{table}[!ht]
\centering
\scriptsize
\setlength{\tabcolsep}{2pt}
\renewcommand{\arraystretch}{1.06}
\begin{tabular}{@{}>{\raggedright\arraybackslash}p{0.17\linewidth}>{\raggedright\arraybackslash}p{0.34\linewidth}>{\raggedright\arraybackslash}p{0.25\linewidth}>{\raggedright\arraybackslash}p{0.14\linewidth}@{}}
\toprule
Marker family & Trigger examples & Code emitted & Fixed before evaluation \\
\midrule
Text reasoning stage & setup, calculate, compute, conclusion, alternative approach & one of setup, calculation, verification, correction, conclusion, exploration, or other & yes \\
Self-verification & check, double-check, verify, confirm, make sure, sanity check & \texttt{sv\_verification} & yes \\
Correction & wait, actually, mistake, wrong, fix, reconsider & \texttt{sv\_correction} & yes \\
Uncertainty & maybe, perhaps, not sure, unclear, might be, could be & \texttt{sv\_uncertainty} & yes \\
Alternative reasoning & alternatively, another way, different approach, try another & \texttt{sv\_alternative} & yes \\
Answer commitment & final answer, answer is, boxed, \texttt{\textbackslash boxed\{\}} & conclusion/final-answer marker & yes \\
No explicit marker & none of the above markers fires & \texttt{sv\_none} & yes \\
\bottomrule
\end{tabular}
\caption{Deterministic text-marker and self-verification rules. The marker coders use lowercase substring matches on the model-generated continuation in the current prefix only, not the original problem prompt. The self-verification family groups correction/verification/uncertainty/commitment markers into interpretable prefix codes; answer-commitment language is represented in the text-marker family through explicit conclusion/final-answer triggers.}
\label{tab:marker-rule-details}
\end{table}

\paragraph{Discriminative prefix-feature baseline.}
Table~\ref{tab:prefix-feature-diagnostic} adds the strongest same-prefix discriminative baseline used in the paper. The classifier is trained only on calibration traces and uses score summaries, observed prefix length, and concept-code features from the same prefix-safe observation stream. This audit leaves the MATH-500 Level 5 text markers as the most direct structure-aware row and keeps GSM8K and RIMO self-verification markers useful, while narrowing the hidden-cluster story on RIMO. Under the split-seed best-variant audit, self-verification retains small positive gaps on both 14B RIMO settings, but paired fixed-hybrid comparisons against PFC are negative or near-tie. RIMO self-verification shows useful prefix information in the observation family; the fixed \method{} readout does not uniformly dominate a flexible prefix classifier.

The PFC is an L2-regularized logistic classifier with fixed optimization settings. Its score features are last-prefix score, EMA, moving average, mean score, score delta, and observed prefix length. Its concept features are the last-prefix concept-code one-hot vector, the concept-code frequency vector over the observed prefix, and a concept transition-rate feature. The classifier is fit on calibration traces only, using the same question-level split as the reported row; no additional probability calibrator is fit on test data. No test-set labels or held-out metrics are used to tune the classifier, so PFC is a same-prefix audit of rank signal, not a separate test-set-selected method.

\begin{table}[!ht]
\centering
\scriptsize
\setlength{\tabcolsep}{2pt}
\renewcommand{\arraystretch}{1.05}
\begin{tabular}{@{}p{0.25\linewidth}p{0.18\linewidth}rrrr@{}}
\toprule
Setting & Observation & Seeds & Gap & Pos. & PFC-best \\
\midrule
MATH-500 Level 5 / 7B & text markers & $50$ & $+0.069$ & $0.98$ & $39/50$ \\
MATH-500 Level 5 / 7B & self-verification & $50$ & $+0.029$ & $0.78$ & $24/50$ \\
MATH-500 Level 5 / 7B & hidden clusters & $50$ & $+0.019$ & $0.70$ & $32/50$ \\
GSM8K / 7B & text markers & $50$ & $+0.031$ & $0.78$ & $25/50$ \\
GSM8K / 7B & self-verification & $50$ & $+0.026$ & $0.76$ & $32/50$ \\
RIMO-N DeepSeek-Qwen-14B & self-verification & $50$ & $+0.023$ & $0.84$ & $11/50$ \\
RIMO-N Qwen3-14B & self-verification & $50$ & $+0.021$ & $0.76$ & $30/50$ \\
RIMO-N DeepSeek-Qwen-14B & hidden clusters & $20$ & $-0.029$ & $0.10$ & $20/20$ \\
RIMO-N Qwen3-14B & hidden clusters & $20$ & $-0.017$ & $0.20$ & $19/20$ \\
AIME 2025 / DeepSeek-Qwen-14B & self-verification & $50$ & $-0.029$ & $0.30$ & $22/50$ \\
Qwen3-14B / MATH-500 Level 5 & self-verification & $50$ & $-0.023$ & $0.28$ & $11/50$ \\
\bottomrule
\end{tabular}
\caption{Prefix-feature classifier (PFC) diagnostic over rewritten text-marker records, rewritten self-verification records, and train-split hidden-cluster records. Gap is the split-seed mean AUROC gap between the best online \method{} variant and the best eligible prefix-safe baseline after adding PFC. Pos. is the fraction of valid split seeds with a positive gap. PFC-best counts seeds in which the classifier is the strongest eligible baseline. Hidden-cluster rows with the larger 64-cluster setting use 20 split seeds because the CPU audit is substantially slower; other rows use 50.}
\label{tab:prefix-feature-diagnostic}
\end{table}

\paragraph{Structure-aware paired diagnostic.}
Table~\ref{tab:temporal-recursion-diagnostic} aligns the split-seed prefix-feature classifier audit with paired bootstrap checks. The classifier column asks whether the same prefix features absorb the rank signal. The emission-only column compares the fixed hybrid \method{} readout with an emission-only hybrid readout that uses the final prefix observations but no recursive Bayesian filtering. The result is a scope check on temporal recurrence: MATH/GSM rows show added rank information, while RIMO rows show absorption by same-prefix evidence.

\medskip
\noindent\begin{minipage}{\linewidth}
\centering
\scriptsize
\setlength{\tabcolsep}{2pt}
\renewcommand{\arraystretch}{1.05}
\begin{tabular}{@{}p{0.22\linewidth}p{0.15\linewidth}rrr@{}}
\toprule
Setting & Observation & Split-seed gap & $\Delta\AUROC$ vs. PFC [95\% CI] & $\Delta\AUROC$ vs. emission-only [95\% CI] \\
\midrule
MATH-500 Level 5 / 7B & text markers & $+0.069$ & $+0.026\;[-0.068,+0.114]$ & $+0.103\;[-0.030,+0.209]$ \\
MATH-500 Level 5 / 7B & self-verification & $+0.029$ & $+0.004\;[-0.070,+0.078]$ & $+0.124\;[-0.027,+0.235]$ \\
MATH-500 Level 5 / 7B & hidden clusters & $+0.019$ & $+0.100\;[+0.023,+0.203]$ & $+0.204\;[-0.009,+0.354]$ \\
GSM8K / 7B & hidden clusters & $-0.005$ & $+0.012\;[-0.055,+0.070]$ & $+0.116\;[+0.045,+0.183]$ \\
RIMO-N DeepSeek-Qwen-14B & self-verification & $+0.023$ & $-0.102\;[-0.167,-0.036]$ & $-0.046\;[-0.131,+0.039]$ \\
RIMO-N DeepSeek-Qwen-14B & hidden clusters & $-0.029$ & $-0.236\;[-0.326,-0.145]$ & $-0.110\;[-0.253,+0.020]$ \\
RIMO-N Qwen3-14B & self-verification & $+0.021$ & $-0.084\;[-0.178,+0.020]$ & $+0.045\;[-0.058,+0.154]$ \\
RIMO-N Qwen3-14B & hidden clusters & $-0.017$ & $-0.048\;[-0.093,-0.008]$ & $+0.044\;[-0.030,+0.117]$ \\
\bottomrule
\end{tabular}
\refstepcounter{table}\label{tab:temporal-recursion-diagnostic}
\par\smallskip
{\small\raggedright\textbf{Table~\thetable.} Structure-aware paired diagnostic from question-cluster bootstrap intervals and split-seed audits. Split-seed gap is the mean AUROC gap after adding PFC to the eligible prefix-safe baselines. The paired PFC column is fixed hybrid \method{} minus the prefix-feature classifier on the same test questions; the emission-only column is fixed hybrid \method{} minus an emission-only hybrid readout using the same final-prefix score and concept observations. The mixed RIMO signs show that the useful RIMO signal can be captured by a same-prefix readout or a flexible prefix classifier.\par}
\end{minipage}

\section{Reliability Diagrams}
\label{sec:reliability-diagnostics}

Table~\ref{tab:reliability-summary} and Figure~\ref{fig:reliability-curves} report reliability diagnostics for representative rows. The curves show that calibrated scalar baselines can be strong probability readouts, while \method{} can still improve Brier or rank in the same setting. The main text treats the identity state readout as a model score unless an additional outcome readout is fitted.

\begin{table}[!ht]
\centering
\scriptsize
\setlength{\tabcolsep}{3pt}
\renewcommand{\arraystretch}{1.06}
\begin{tabular}{@{}p{0.27\linewidth}p{0.30\linewidth}p{0.35\linewidth}@{}}
\toprule
Setting & Reliability pattern & Diagnostic takeaway \\
\midrule
MATH-500 Level 5 / 7B & Structure-aware observations improve rank, but calibrated scalar baselines remain competitive as probability readouts. & Calibration-limit evidence. \\
GSM8K / 7B & Hybrid \method{} gives an easier-regime reliability curve with useful Brier and rank behavior. & Representative easier-regime reliability curve. \\
RIMO / AIME stress tests & Identity-readout ECE is setting-dependent even when Brier or AUROC are informative. & Caution against probability-readout overclaim. \\
\bottomrule
\end{tabular}
\caption{Reliability-diagram summary for representative rows.}
\label{tab:reliability-summary}
\end{table}

\noindent\begin{minipage}{\linewidth}
 \centering
 \includegraphics[width=0.48\linewidth]{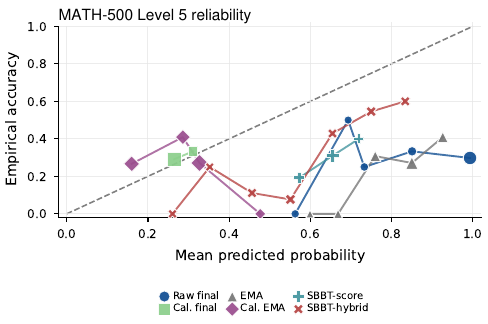}
 \includegraphics[width=0.48\linewidth]{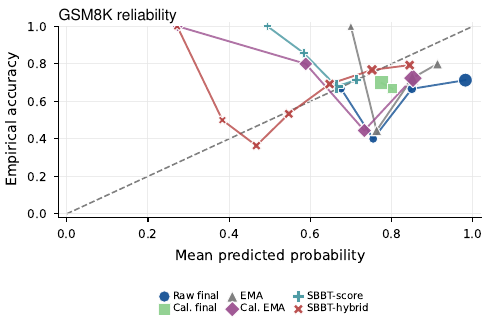}
 \refstepcounter{figure}\label{fig:reliability-curves}
 \par\smallskip
 {\small\raggedright\textbf{Figure~\thefigure.} Reliability curves for representative MATH-500 Level 5 and GSM8K rows. The curves compare raw last-prefix score, calibrated last-prefix score, EMA, calibrated EMA, score-only \method{}, and hybrid \method{} readouts. They support the calibration/ranking decoupling analysis and motivate treating identity state readout as a model score unless an outcome readout is fitted.\par}
\end{minipage}

\section{Additional Results}

Table~\ref{tab:rollout-validation}, Figure~\ref{fig:rollout-belief-validation}, Figure~\ref{fig:utility}, Table~\ref{tab:answer-audit-outcomes}, and Table~\ref{tab:model-family-diagnostic} report completed rollout joins, utility operating points, answer-label audit outcomes, and model-family diagnostics. The DeepSeek-Llama-8B row checks whether the MATH-500 Level 5 observation-family pattern survives outside the main DeepSeek-Qwen row. Scoped token-pooling diagnostics remain stress-test or near-tie diagnostics. Figure~\ref{fig:rimo-diagnostics} isolates the layerwise hidden-cluster pattern: train-split hidden-cluster observations are positive at early, mid, and final layers, while direct hidden probes and activation trajectories do not become main rows.

\noindent\begin{minipage}{\linewidth}
\centering
\scriptsize
\setlength{\tabcolsep}{2pt}
\begin{tabular}{@{}p{0.21\linewidth}p{0.18\linewidth}p{0.24\linewidth}p{0.29\linewidth}@{}}
\toprule
Dataset/model & Coverage & Source score & \method{} belief summary \\
\midrule
RIMO-N DeepSeek-Qwen-14B & $120$ groups / $480$ continuations; mean success $0.3312$ & Brier $0.5321$; Pearson $-0.0067$; Spearman $0.1070$ & Brier $0.2473$; Pearson $-0.1125$; Spearman $-0.1081$. Calibration gain only; rank remains weak. \\
RIMO-N Qwen3-14B & $120$ groups / $480$ continuations; mean success $0.3938$ & Brier $0.4623$; Pearson $0.0127$; Spearman $0.0881$ & Brier $0.2151$; Pearson $0.0730$; Spearman $0.0929$. Strong Brier improvement; rank remains weak. \\
MATH-500 Level 5 DeepSeek-Qwen-7B & $120$ groups / $480$ continuations; mean success $0.3729$ & Brier $0.2964$; Pearson $0.2150$; Spearman $0.2116$ & Brier $0.2192$; Pearson $0.1495$; Spearman $0.1441$. Supports probability-quality transfer; source score keeps stronger rank correlation. \\
\bottomrule
\end{tabular}
\refstepcounter{table}\label{tab:rollout-validation}
\par\smallskip
\parbox{0.94\linewidth}{\small\textbf{Table~\thetable. Rollout-based prefix-value validation on held-out test prefixes.}
The completed belief joins strengthen the paper's calibration/ranking decoupling claim: \method{} beliefs are closer to empirical continuation-success probabilities in Brier score on all three held-out prefix sets, while rank association is setting-dependent and does not consistently dominate the raw source score. The rollout estimate is empirical continuation success under the chosen rollout policy and sampling protocol, with approximate binomial uncertainty $\mathrm{SE}(\hat v_t)\approx\sqrt{\hat v_t(1-\hat v_t)/K}$; this standard error treats sampled continuations as conditionally independent under the rollout protocol.}
\end{minipage}

\noindent\begin{minipage}{\linewidth}
 \centering
 \includegraphics[width=0.90\linewidth]{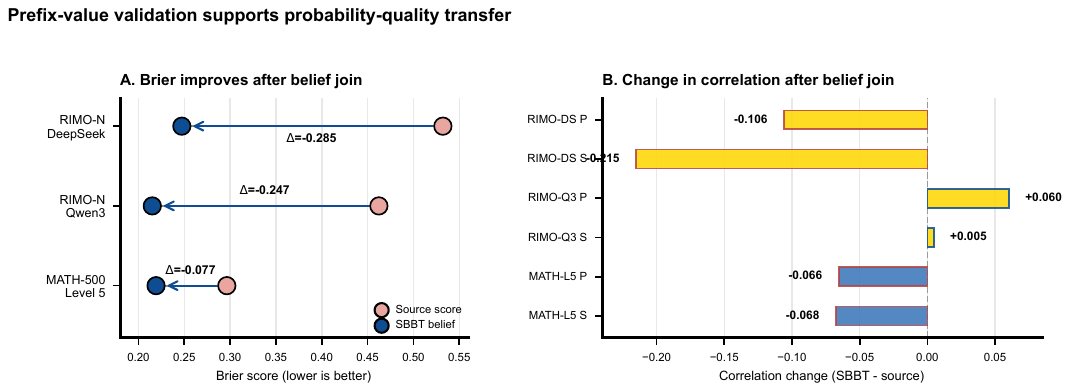}
 \refstepcounter{figure}\label{fig:rollout-belief-validation}
 \parbox{0.94\linewidth}{\small\textbf{Figure~\thefigure. Rollout-based calibration diagnostic.}
 Completed RIMO-N DeepSeek, RIMO-N Qwen3, and MATH-500 Level 5 belief joins compare raw source scores and \method{} beliefs against empirical continuation success. Panel A shows Brier improvement after the belief join. Panel B, Change in correlation after belief join, reports $\Delta$ correlation = \method{} belief minus source score for Pearson and Spearman correlations. Together, the panels show probability-quality transfer with setting-dependent rank association.}
\end{minipage}

\medskip
\noindent\begin{minipage}{\linewidth}
 \centering
 \includegraphics[width=0.62\linewidth]{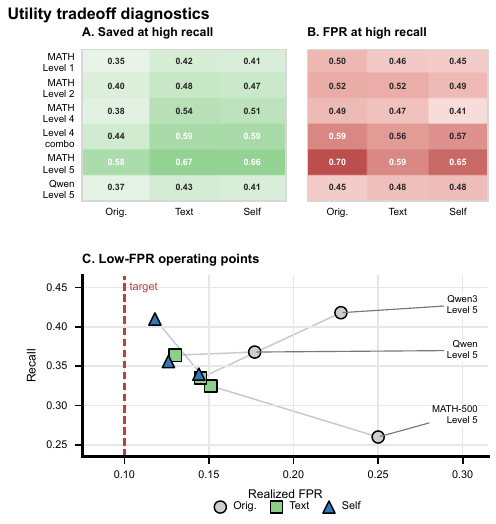}
 \refstepcounter{figure}\label{fig:utility}
 \par\smallskip
 \parbox{0.72\linewidth}{\small\textbf{Figure~\thefigure. Utility operating points from existing split-seed summaries.}
 Panels A/B summarize high-recall operating points as saved bad-trace compute and false-positive rate, respectively. Panel C shows realized FPR and recall for low-FPR operating points. Text and self-verification observations can improve savings, but false positives remain high enough that the false-positive rates motivate decision-oriented evaluation without deployment claims.}
\end{minipage}

\noindent\begin{minipage}{\linewidth}
\centering
\scriptsize
\setlength{\tabcolsep}{2pt}
\renewcommand{\arraystretch}{1.05}
\begin{tabular}{@{}p{0.25\linewidth}rrrrrp{0.24\linewidth}@{}}
\toprule
Row & Reviewed samples & Missing & Cap hits & Possible equiv. & Label changes & Scope \\
\midrule
MATH-500 Level 5 / 7B & 54 & 0 & 0 & 2 & 0 & Symbolic-equivalence review. \\
AIME 2025 / 14B & 57 & 0 & 0 & 0 & 0 & Boxed-answer extraction check. \\
RIMO-N / DeepSeek-Qwen-14B & 76 & 0 & 0 & 0 & 0 & Broad hard-benchmark label review. \\
RIMO-N / Qwen3-14B & 78 & 0 & 0 & 1 & 0 & Prompt-protocol label review. \\
\bottomrule
\end{tabular}
\refstepcounter{table}\label{tab:answer-audit-outcomes}
\par\smallskip
\parbox{0.94\linewidth}{\small\textbf{Table~\thetable. Manual answer-label audit outcomes for reported hard-math rows.}
Reviewed samples were selected from missing-answer, token-cap, and possible-equivalence categories identified by the automatic sampler. This is a risk-focused audit over automatically flagged cases, not a full random relabeling study. Manual review found no label changes for the reported rows.}
\end{minipage}

\medskip
\noindent\begin{minipage}{\linewidth}
\centering
\scriptsize
\setlength{\tabcolsep}{3pt}
\renewcommand{\arraystretch}{1.05}
\begin{tabular}{@{}p{0.30\linewidth}rrrrrrr@{}}
\toprule
Dataset/model & Err. & Obs. & Score & Text & Self & HC & LT \\
\midrule
DeepSeek-Llama-8B / MATH-500 Level 5 & 78.2 & 39.2 & +0.004 & +0.030 & +0.017 & +0.058 & \textbf{+0.094} \\
\bottomrule
\end{tabular}
\refstepcounter{table}\label{tab:model-family-diagnostic}
\par\smallskip
{\small\raggedright\textbf{Table~\thetable.} Model-family diagnostic on the completed DeepSeek-R1-Distill-Llama-8B MATH-500 Level 5 full run. The row uses $134$ questions and $536$ traces with four traces per question. Entries report split-seed mean AUROC gaps against the strongest eligible prefix-safe baseline. The result is kept as appendix evidence: structure-aware observations remain positive outside the main DeepSeek-Qwen row, while score dynamics are not a positive signal.\par}
\end{minipage}

\medskip
\noindent\begin{minipage}{\linewidth}
 \centering
 \includegraphics[width=0.92\linewidth]{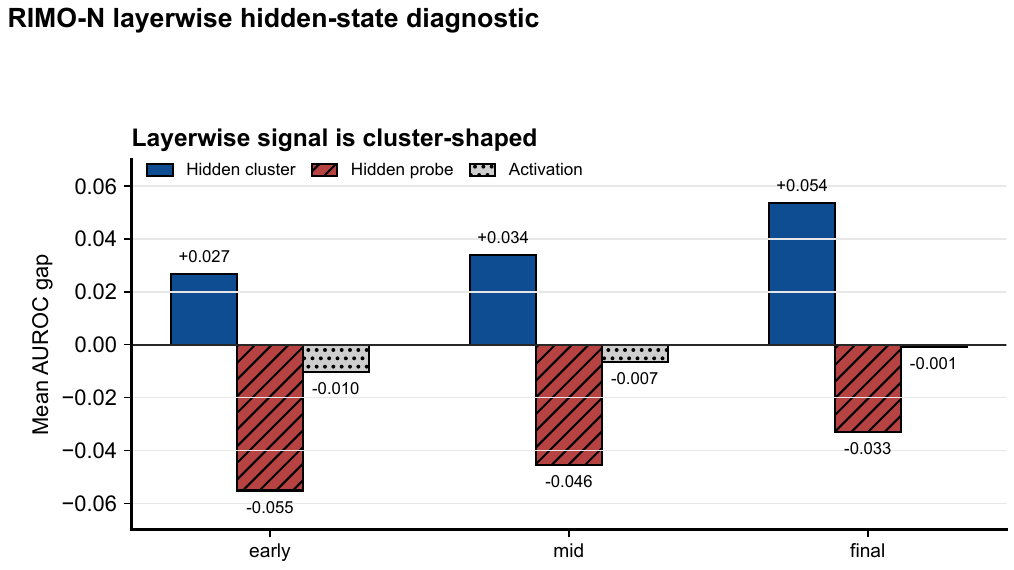}
 \refstepcounter{figure}\label{fig:rimo-diagnostics}
 \par\smallskip
 {\small\raggedright\textbf{Figure~\thefigure.} Layerwise RIMO outcomes. Train-split hidden clustering is positive at early, mid, and final layers, while direct hidden probes and activation trajectories remain negative or near-tie. This separates a useful RIMO hidden-state signal from two weaker hidden-state variants under the same split-seed protocol.\par}
\end{minipage}

Figures~\ref{fig:split-seed-distribution}, \ref{fig:evidence-transfer}, \ref{fig:prefix-trajectory-ribbon}, and \ref{fig:baseline-dominance} collect the appendix evidence for the paper's multi-view argument: split-seed robustness, observation-family transfer, prefix-trajectory behavior, and baseline-strength effects.

\medskip
\noindent\begin{minipage}{\linewidth}
 \centering
 \includegraphics[width=0.96\linewidth]{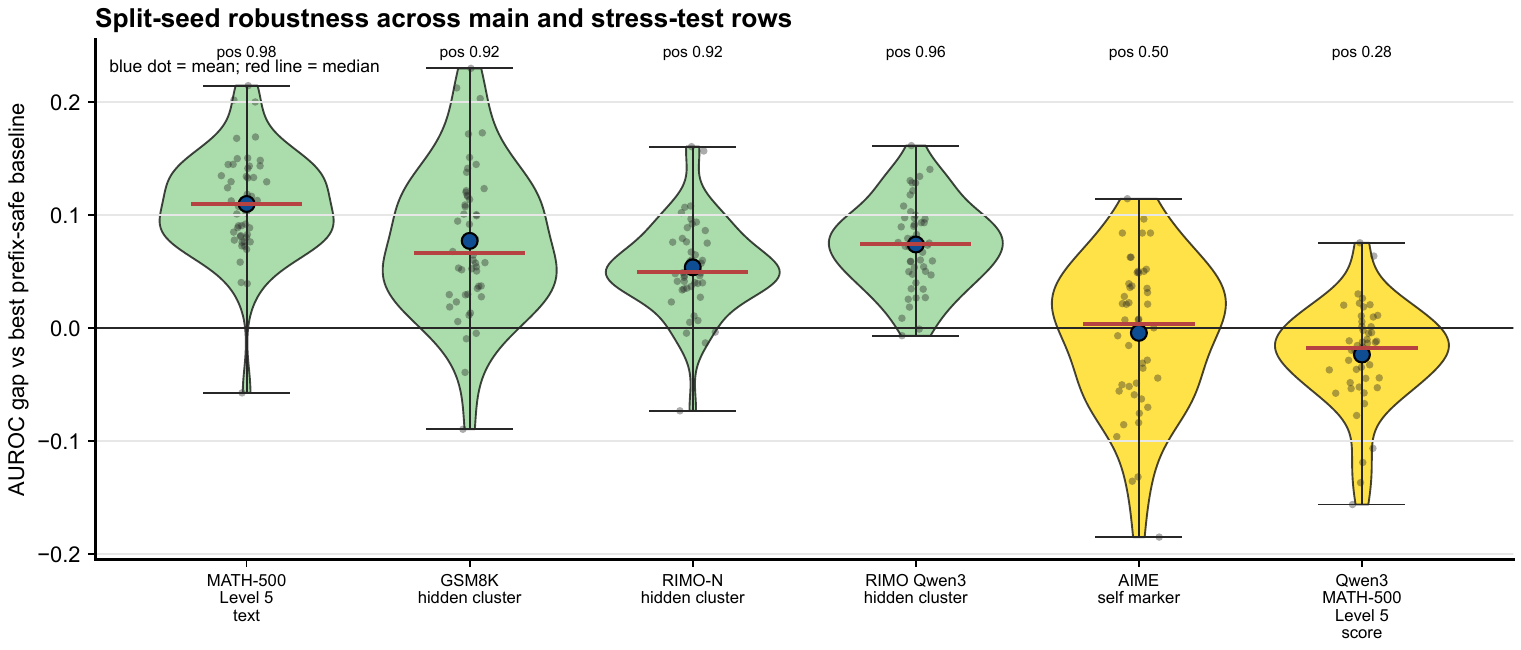}
 \refstepcounter{figure}\label{fig:split-seed-distribution}
 \par\smallskip
 {\small\raggedright\textbf{Figure~\thefigure.} Split-seed AUROC-gap distributions for representative main and stress-test rows. In the pre-PFC split-seed view, structure-aware rows are positive across most question-level split seeds for MATH-500, GSM8K, and the RIMO observation families shown here; AIME self-verification and Qwen3-MATH score rows sit near or below zero, which motivates keeping them as stress-test rows.\par}
\end{minipage}

\medskip
\noindent\begin{minipage}{\linewidth}
 \centering
 \includegraphics[width=0.96\linewidth]{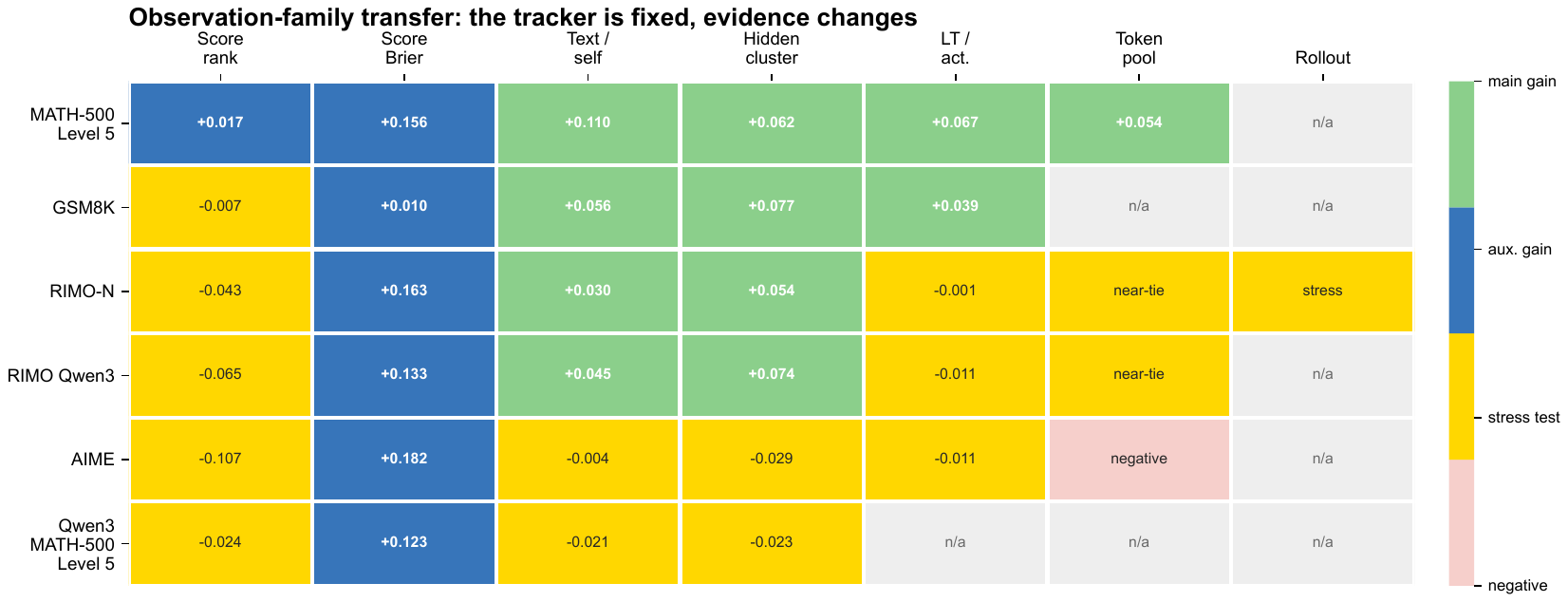}
 \refstepcounter{figure}\label{fig:evidence-transfer}
 \par\smallskip
 {\small\raggedright\textbf{Figure~\thefigure.} Observation-family transfer matrix. Cells summarize main gains, auxiliary gains, stress-test rows, and not-applicable entries.\par}
\end{minipage}

\medskip
\noindent\begin{minipage}{\linewidth}
 \centering
 \includegraphics[width=0.94\linewidth]{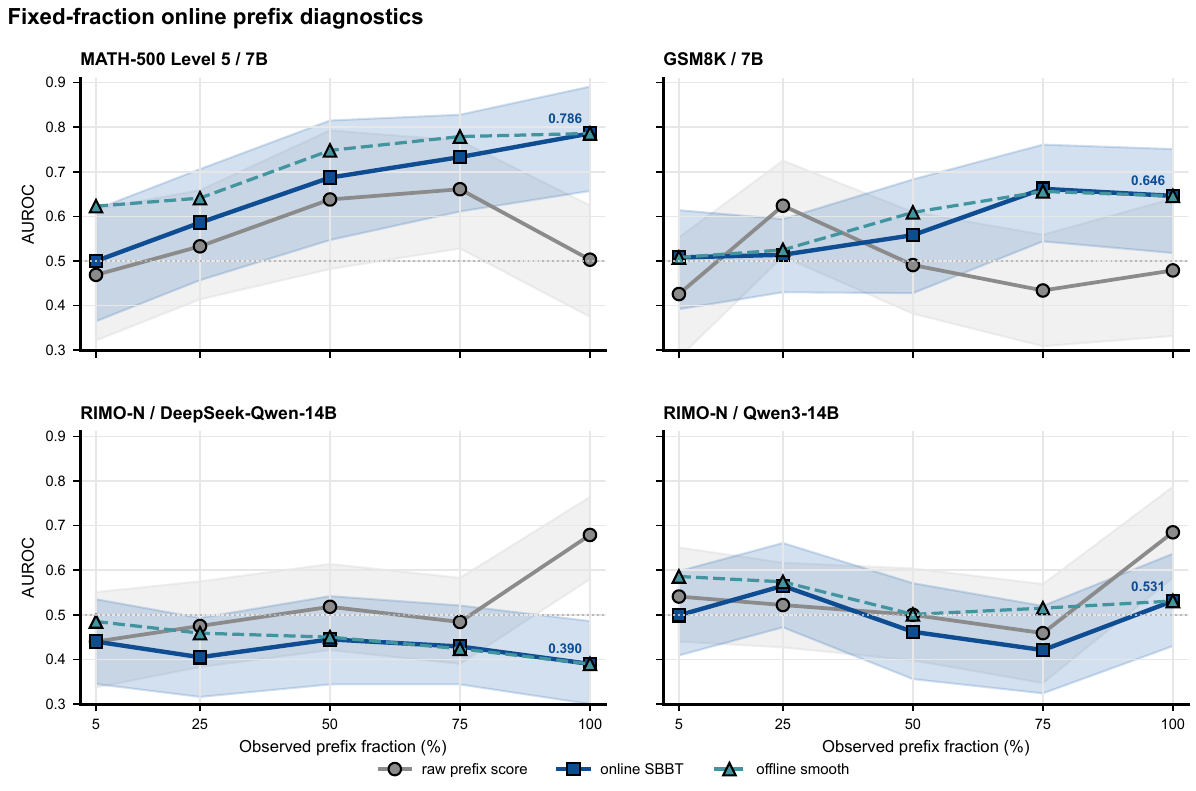}
 \refstepcounter{figure}\label{fig:prefix-trajectory-ribbon}
 \par\smallskip
 {\small\raggedright\textbf{Figure~\thefigure.} Fixed-fraction prefix diagnostics with question-cluster bootstrap intervals. Available exports provide 5\%, 25\%, 50\%, 75\%, and 100\% prefix fractions for MATH-500 Level 5, GSM8K, and two RIMO-N rows. MATH-500 Level 5 and GSM8K show late-prefix online gains, while the RIMO online hybrid curves remain stress-test rows despite final hidden-cluster and self-verification gains.\par}
\end{minipage}

\medskip
\noindent\begin{minipage}{\linewidth}
 \centering
 \includegraphics[width=0.90\linewidth]{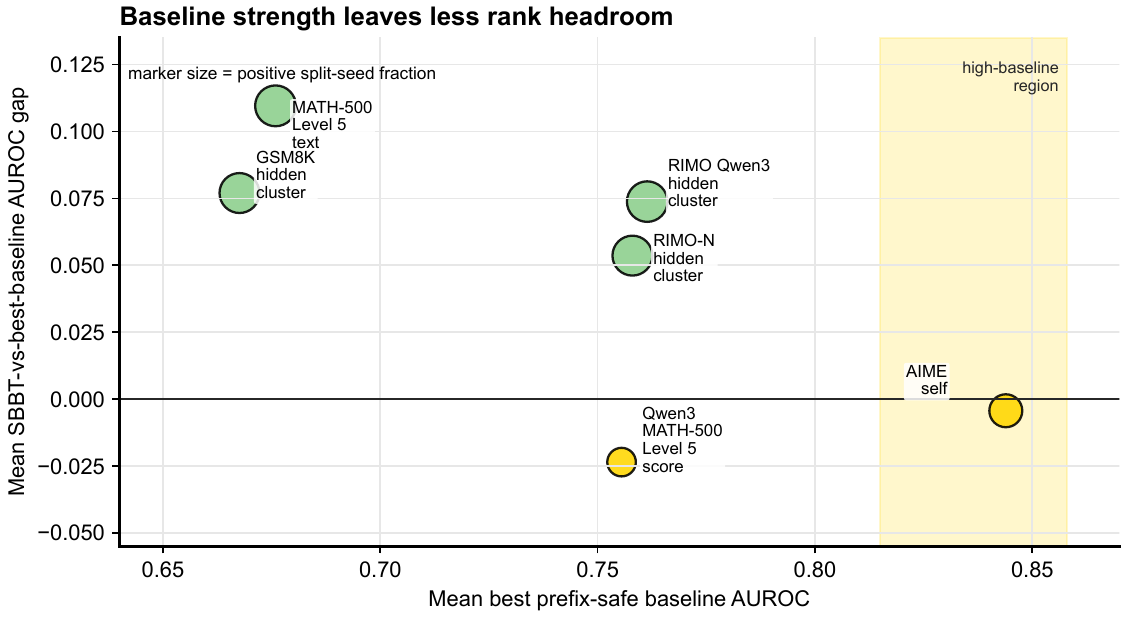}
 \refstepcounter{figure}\label{fig:baseline-dominance}
 \par\smallskip
 {\small\raggedright\textbf{Figure~\thefigure.} Prefix-safe baseline headroom diagnostic. Rows with high best-baseline AUROC leave less rank headroom; RIMO hidden-cluster observations improve rank against standard prefix-safe baselines before adding the same-prefix PFC audit, while AIME and Qwen3-MATH remain stress-test cases. Marker size encodes positive split-seed fraction.\par}
\end{minipage}

\section{Signed AUROC Gaps and Evidence Coverage}
\label{sec:appendix-diagnostics}

Figure~\ref{fig:signed-auroc-boundary} reports the signed AUROC gaps that are intentionally not used as the main visual summary. The negative values are important stress-test evidence: they show that score-only, token-pooling, AIME-style, and some Qwen-family MATH settings can be dominated by strong prefix-safe baselines. Figure~\ref{fig:diagnostic-coverage} records which result families have rank, probability, early-prefix, utility, rollout, and audit evidence. The main text uses Figure~\ref{fig:gaps} for the compact summary and keeps these appendix plots as reference diagnostics.

\medskip
\noindent\begin{minipage}{\linewidth}
 \centering
 \includegraphics[width=0.96\linewidth]{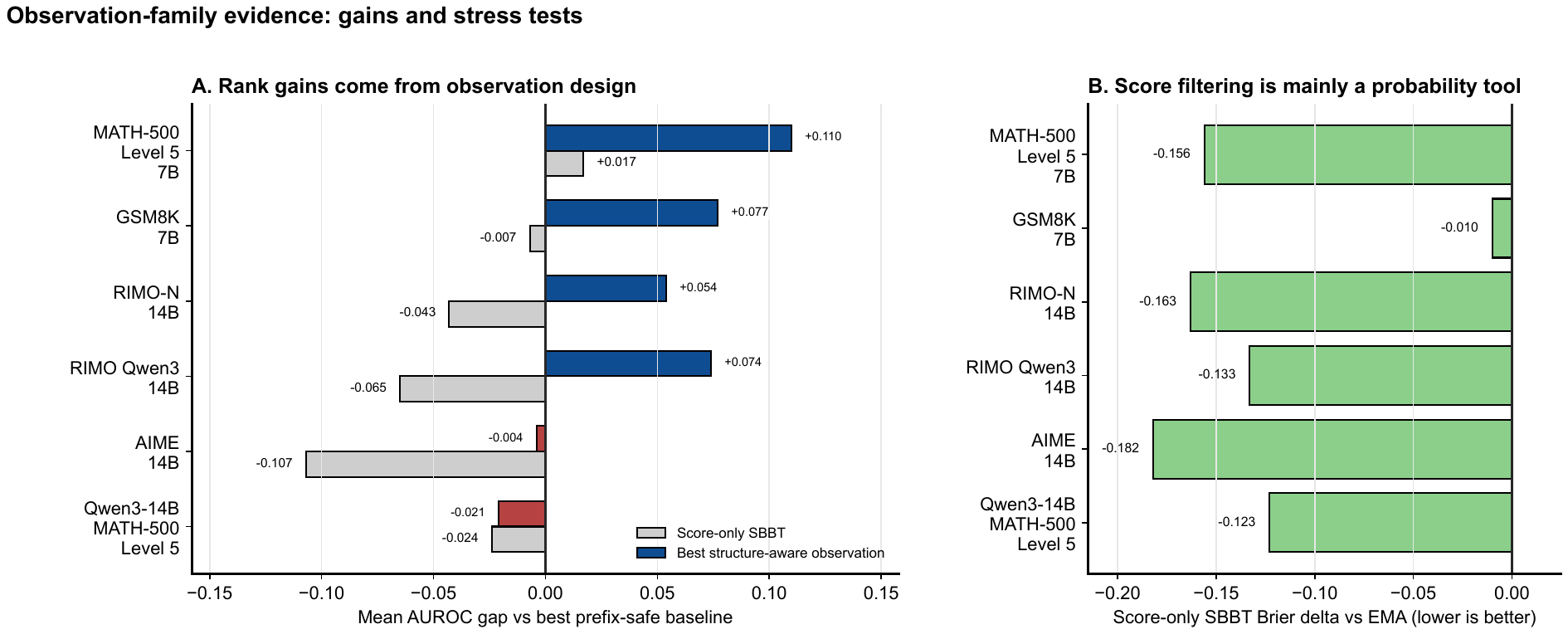}
 \refstepcounter{figure}\label{fig:signed-auroc-boundary}
 \par\smallskip
 {\small\raggedright\textbf{Figure~\thefigure.} Signed observation-family AUROC gaps and score-only Brier deltas. Positive AUROC gaps indicate a structure-aware observation-family gain; negative AUROC gaps mark settings where the available observation family does not beat the strongest prefix-safe baseline.\par}
\end{minipage}

\medskip
\noindent\begin{minipage}{\linewidth}
 \centering
 \includegraphics[width=0.96\linewidth]{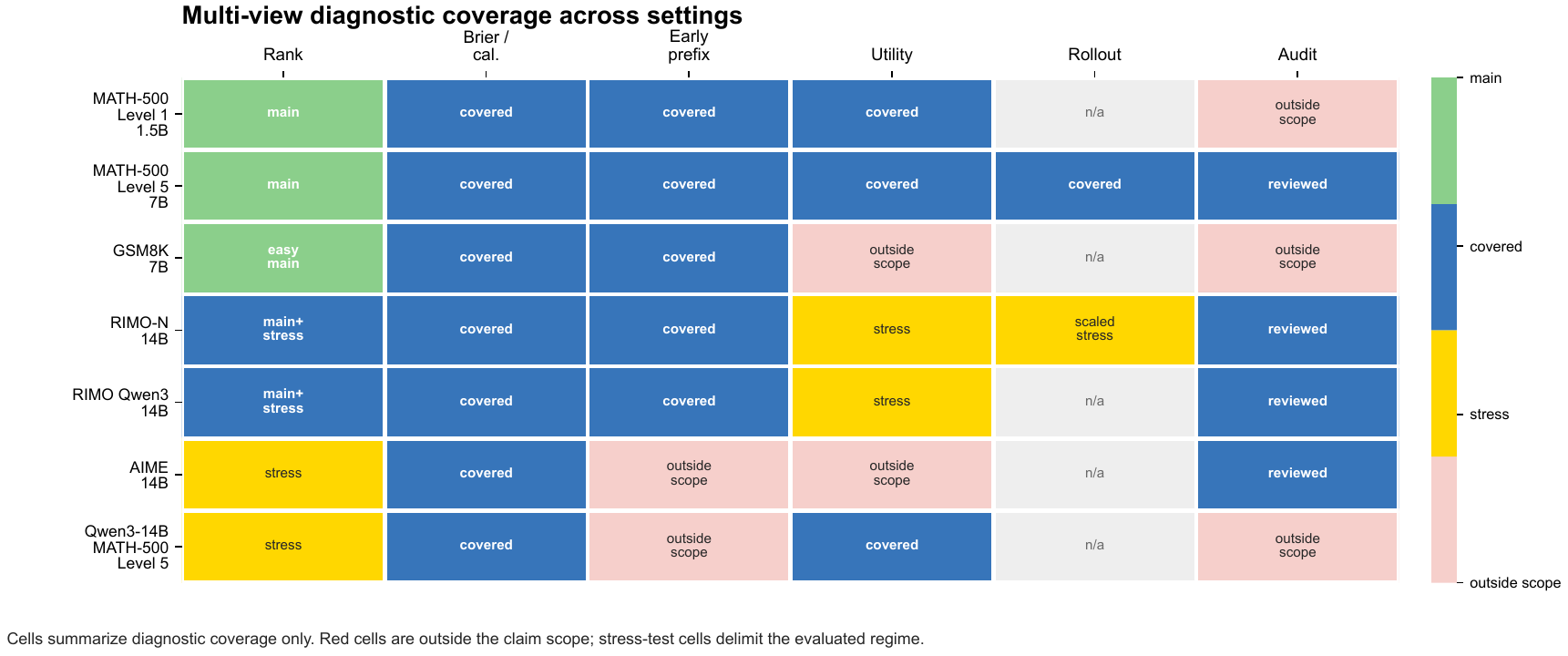}
 \refstepcounter{figure}\label{fig:diagnostic-coverage}
 \par\smallskip
 {\small\raggedright\textbf{Figure~\thefigure.} Multi-view diagnostic coverage across evaluated settings. Rows summarize which dataset/model settings have rank, probability, early-prefix, utility, rollout, and answer-audit evidence. Cells encode diagnostic coverage only.\par}
\end{minipage}

\section{Reproducibility Details}
\label{sec:reproducibility}

This appendix records reproducibility details for the reported rows. Exact commands, local paths, and environment setup are not part of the paper text; the paper records the protocol and artifact classes needed to audit the results. An anonymized software artifact is submitted with the paper, and we will release a public non-anonymous repository after review. Table~\ref{tab:repro-artifacts} summarizes the reproducibility record. The primary empirical path uses generated open-weight traces: prefix observations, calibration, filtering, split-seed sweeps, diagnostics, and figures can be regenerated once benchmark inputs, model checkpoints, and trace files are available.

\subsection{Reproducibility Record}

\begin{center}
\small
\setlength{\tabcolsep}{3pt}
\renewcommand{\arraystretch}{1.06}
\begin{tabular}{p{0.20\linewidth}p{0.74\linewidth}}
\toprule
Component & Reproduction detail \\
\midrule
Code and environment & The submitted anonymized software artifact covers generation, feature extraction, calibration, filtering, diagnostics, plotting, and paper-refresh code. Dependency versions, hardware notes, and executable entry points are recorded in reproducibility metadata. \\
Paper assets & The paper source and rendered figures are maintained with the submission draft. Exact shell commands and non-anonymous repository URLs are not part of the paper text. \\
Benchmark inputs & GSM8K, MATH-500, AIME 2025, and RIMO-N inputs are normalized into local JSONL files before batch generation runs. JSONL fields must include a question id, problem text, gold answer, and optional difficulty/level field. \\
Models & Paper rows use open-weight checkpoints listed in the reproducibility manifest, including DeepSeek-R1-Distill-Qwen, DeepSeek-R1-Distill-Llama, Qwen2.5-Math, Qwen3, and DeepSeek-Qwen variants. The main text cites model-family reports; exact checkpoint identifiers and optional model-card entries belong to the reproducibility metadata and bibliography backup. \\
Licenses and intended use & Reproducibility metadata records the source, license, and access terms for benchmark inputs, checkpoints, and generated artifacts. Derived traces and diagnostics are intended for research auditing of reasoning-reliability methods and inherit the access constraints of the original artifacts. \\
PII and content scope & The inputs are public math-reasoning benchmarks, not newly collected personal data. Diagnostic summaries omit author/user identifiers; generated traces are used for research inspection under the artifact terms recorded in the manifest. \\
Compute scale & Reported rows use open-weight checkpoints in the 1.5B--14B parameter range on A100-class GPUs. The end-to-end experimental cycle used approximately 700 A100 GPU-hours, including exploratory runs, protocol-correction reruns, open-weight trace generation, rollout continuations, and post-generation diagnostics. Run metadata records checkpoint identifiers, shard counts, decoding settings, wall-clock notes, and available GPU-hour accounting for batch generation and diagnostics. \\
Generated traces and diagnostics & Exact reproduction requires the underlying \texttt{records.jsonl}, feature JSONL files, run metadata, generation manifests, split definitions, and diagnostic summaries for each paper row. The artifact manifest records file sizes and SHA256 checksums for generated traces, feature files, split definitions, and diagnostic summaries. \\
Answer labels & Final-answer labels are produced by dataset-specific answer extraction/checking, with symbolic checking used only where explicitly configured. Manual answer-audit samples did not require label changes for the reported rows; reviewed samples are tracked in the audit record. \\
\bottomrule
\end{tabular}
\par\smallskip
\refstepcounter{table}\label{tab:repro-artifacts}
{\small\textbf{Table~\thetable.} Reproducibility record for the reported rows. Generated data are intentionally kept separate from the submission source.}
\end{center}

\subsection{Evaluation Protocol}

All diagnostic rows use question-level train/calibration/test partitions: traces from the same original problem never cross partitions. Observation-family parameters that can leak information, including hidden probes, score orientations, hidden-cluster assignments, calibration maps, emission likelihoods, utility thresholds, learned-prefix baselines, and prefix-feature classifiers, are fitted only on train/calibration questions and then applied to held-out test questions. Split-seed robustness sweeps repeat this protocol across question-level seeds; AUROC is reported only on valid held-out splits containing both final-success classes. Brier deltas are reported as method minus baseline, so negative deltas mean better probability quality.

The default online score is the final prefix belief $\pi_T(\stateon)$ for each trace. Early-prefix figures separately evaluate fixed prefix fractions such as 5\%, 25\%, 50\%, 75\%, and 100\%. Forward-backward smoothing, Viterbi paths, and rollout-based prefix-value joins are offline checks and are not used for online-intervention claims.

\subsection{Artifact Contents and Variability}

A complete reproduction record includes the submitted software artifact, implementation metadata, run manifests, normalized benchmark inputs or source identifiers, model identifiers, generated traces, feature files, split definitions, diagnostic summaries, answer-audit records, license/terms metadata, compute-scale notes, and checksum manifest. Experiments were run on A100-class GPUs using cached benchmark JSONL files and model checkpoints. GPU count changes wall-clock time, while the statistical protocol is fixed by question-level splits, trace indices, shard offsets, decoding settings, and random seeds recorded in the manifests.

The remaining reproducibility risks are standard for stochastic LLM reasoning experiments. Exact trace-level reproduction depends on pinned model checkpoints, tokenizer versions, chat templates, recorded seeds, model revisions, run manifests, and checksums. Statistical reproduction reports split-seed, paired-delta, and calibration summaries and does not depend on a single deterministic rerun.

\section{Terminology Glossary}

Table~\ref{tab:notation} in Appendix~\ref{sec:math-details} defines the mathematical symbols used in the derivations. This glossary fixes prose-facing vocabulary and evidence scope in Table~\ref{tab:glossary}. The main consistency rule is that \method{} is the method name used in prose and that reliability states refer to eventual-success evidence, not verified step correctness.

\begin{center}
\scriptsize
\setlength{\tabcolsep}{3pt}
\renewcommand{\arraystretch}{0.96}
\begin{tabular}{p{0.19\linewidth}p{0.75\linewidth}}
\toprule
Term & Paper usage \\
\midrule
\method{} / Sequential Bayesian Belief Tracking & Method name used in prose for the calibration-aware sequential belief tracker. Use this as the method name in claims and captions. \\
Trace & One sampled reasoning trajectory for one question, including generated text, final answer, final correctness, and ordered observations. \\
Prefix & The part of a generated trace available up to time or chunk $t$. Online claims may use only prefixes, never future chunks. \\
Observation & A prefix-safe signal extracted from a generated prefix: score, concept code, hidden-cluster id, text and self-verification marker, probe score, token-pooling score, or latent-trajectory score. \\
Observation family & A class of prefix-safe extraction functions, such as score-only filtering, text and self-verification markers, hidden-cluster observations, hidden probes, token pooling, or latent trajectory. \\
Reliability state & Latent two-state variable used by the filter. It represents eventual-success reliability under the model, not externally verified step correctness. \\
\method{} readout & The online reliability score used in experiments. Optional calibrated readouts can map it to a probability estimate, but the default readout is a modeling convention, not a theorem that the latent state equals the final label. \\
Outcome readout & A calibration map from latent-state posterior beliefs to an eventual-success probability estimate. \\
High-reliability state & Latent state denoted by $\stateon$ in equations: the prefix observation is more likely under traces that eventually answer correctly. \\
Low-reliability state & Latent state denoted by $\stateoff$ in equations: the prefix observation is more likely under traces that eventually fail. \\
Eventual success & Final-answer correctness of the whole generated trace. This is the supervised target in the experiments. \\
Step correctness & Human or process-label correctness of an intermediate reasoning step. This is not provided by final-answer labels. \\
Prefix value & Empirical recoverability of a prefix, estimated by sampled continuations and final-answer scoring. It validates eventual-success reliability, not first-error localization. \\
Stress-test row & A clean setting where current observations do not beat strong prefix-safe baselines. Stress-test rows are used to limit claims, not hidden as failures. \\
Structure-aware observation & Text marker, self-verification marker, hidden-cluster, hidden-probe, token-pooling, or trajectory-derived observation that is intended to expose reasoning structure beyond a scalar score. \\
Scoped token-pooling probe & A token-pooling diagnostic that restricts pooled hidden vectors to a prefix-safe token span such as generated-only or verification-span tokens. RIMO scoped pooling remains negative or near-tie. \\
Probe trajectory & A sequence of probe outputs over generated prefixes or tokens. It is related work and a future observation-family study here, not the same as raw hidden-state mean/max pooling. \\
Emission likelihood & Calibrated observation likelihood used by the Bayesian filter. \\
Transition matrix & Two-state dynamics with error and recovery probabilities. \\
All-prefix calibration & Weak-supervision mode that propagates final trace correctness to all prefixes for likelihood fitting. Use only as eventual-success supervision. \\
Final-step calibration & Calibration mode fitting emissions only on the final observation of each trace. \\
EM calibration & Baum-Welch-style likelihood fitting without directly treating every prefix label as observed. \\
Score-only SBBT & Filter using only a continuous score likelihood, such as entropy, hidden-probe score, or latent-trajectory score. \\
Concept SBBT & Filter using discrete concept-code likelihoods, such as text markers or hidden-cluster observations. \\
Hybrid SBBT & Filter multiplying continuous score and concept likelihoods under a conditional-independence approximation. \\
Joint SBBT & Filter estimating a joint likelihood over score bins and concept codes. \\
Prefix-safe baseline & Baseline whose prediction at $t$ uses only prefix information and calibration/train data. \\
Learned-prefix baseline & Deterministic prefix-only baseline trained on calibration traces using score and length features. It is a strong comparator, not part of SBBT. \\
Prefix-feature classifier & Deterministic discriminative baseline trained on calibration traces using score summaries, observed prefix length, and concept-code features from the same prefix-safe observations. It tests whether a flexible classifier can absorb the evidence used by the tracker. \\
Length-aware baseline & Prefix-safe baseline that can exploit elapsed reasoning length or observed-prefix count. It is especially strong on RIMO/AIME-style traces where shorter traces are often correct. \\
Smoothing / Viterbi & Offline checks that use the full observation sequence. Do not describe them as online methods. \\
AUROC gap & Difference between an SBBT variant and the best baseline under the same split: the standard baseline set unless a PFC audit table explicitly says otherwise. Positive values favor SBBT. \\
Brier delta & Difference in Brier score against a baseline. Negative values indicate better probability quality. \\
MATH-500 Level & Dataset-provided difficulty annotation inherited from the MATH benchmark. Level labels identify dataset slices, not SBBT variants, neural layers, or training stages. \\
Score/rollout correlation & Correlation between a raw prefix score and empirical prefix value. Near-zero values mean the score is not a useful recoverability estimate, even if other observation families remain useful. \\
Utility evaluation & Decision-facing triage evaluation of early bad-trace detection, compute saved, and false positives without formal risk-control guarantees. \\
\bottomrule
\end{tabular}
\par\smallskip
\refstepcounter{table}\label{tab:glossary}
{\small\textbf{Table~\thetable.} Prose-facing terminology used throughout the paper. Mathematical symbols are defined separately in Table~\ref{tab:notation}.}
\end{center}

\end{document}